\documentclass[10pt,journal,compsoc]{IEEEtran}
\usepackage{ifpdf}

\ifCLASSOPTIONcompsoc
  \usepackage[nocompress]{cite}
\else
  \usepackage{cite}
\fi
\ifCLASSINFOpdf
  \usepackage[pdftex]{graphicx}
  \usepackage{epstopdf}
\else
\fi
\usepackage{algorithmic}

\usepackage{array}
\usepackage{color}
\usepackage{algorithm}
\usepackage{algorithmic}
\usepackage{bm}
\usepackage{epsfig}
\usepackage{graphicx}
\usepackage{amsmath}
\usepackage{amssymb}
\usepackage{multirow}
\usepackage{booktabs} 
\usepackage{algorithm}
\usepackage{algorithmic}

\usepackage{adjustbox}
\usepackage{booktabs}
\usepackage{multirow}
\usepackage{subcaption}
\usepackage{tabularx}
\usepackage{float}

\usepackage{url}

\newcommand{\keze}[1]{{\color{black}#1}}
\newcommand{\kezen}[1]{{\color{black}#1}}

\usepackage{amsmath}

\DeclareMathOperator*{\argmin}{arg\,min}

\usepackage{expl3}
\ExplSyntaxOn
\newcommand\latinabbrev[1]{
  \peek_meaning:NTF . {
    #1\@}%
  { \peek_catcode:NTF a {
      #1.\@ }%
    {#1.\@}}}
\ExplSyntaxOff

\def\eg{\latinabbrev{e.g}}
\def\etal{\latinabbrev{et al}}

\def\ie{\latinabbrev{i.e}}

\begin{document}
%
\title{3D Human Pose Machines with \\ Self-supervised Learning}
%
%
%
%

\author{Keze Wang,~
		Liang Lin,~
		Chenhan Jiang,~
        Chen Qian,~
        and Pengxu Wei
\thanks{K. Wang is with the School of Data and Computer Science, Sun Yat-sen University, Guangzhou, China, and also with the Department of Computing, The Hong Kong Polytechnic University, Hong Kong (e-mail: kezewang@gmail.com).}
\thanks{L. Lin, P. Wei, and C. Jiang are with the School of Data and Computer Science, Sun Yat-sen University, Guangzhou, China. L. Lin is the corresponding author; e-mail: linlang@ieee.org}
\thanks{C. Qian is with SenseTime Group.}
}
\markboth{IEEE TRANSACTIONS ON PATTERN ANALYSIS AND MACHINE INTELLIGENCE, 2019.}%
{}
%



\IEEEtitleabstractindextext{
\begin{abstract}
Driven by recent computer vision and robotic applications, recovering 3D human poses has become increasingly important and attracted growing interests. In fact, completing this task is quite challenging due to the diverse appearances, viewpoints, occlusions and inherently geometric ambiguities inside monocular images. Most of the existing methods focus on designing some elaborate priors /constraints to directly regress 3D human poses based on the corresponding 2D human pose-aware features or 2D pose predictions. However, due to the insufficient 3D pose data for training and the domain gap between 2D space and 3D space, these methods have limited scalabilities for all practical scenarios (e.g., outdoor scene). Attempt to address this issue, this paper proposes a simple yet effective self-supervised correction mechanism to learn all intrinsic structures of human poses from abundant images. Specifically, the proposed mechanism involves two dual learning tasks, i.e., the 2D-to-3D pose transformation and 3D-to-2D pose projection, to serve as a bridge between 3D and 2D human poses in a type of ``free'' self-supervision for accurate 3D human pose estimation. The 2D-to-3D pose implies to sequentially regress intermediate 3D poses by transforming the pose representation from the 2D domain to the 3D domain under the sequence-dependent temporal context, while the 3D-to-2D pose projection contributes to refining the intermediate 3D poses by maintaining geometric consistency between the 2D projections of 3D poses and the estimated 2D poses. Therefore, these two dual learning tasks enable our model to adaptively learn from 3D human pose data and external large-scale 2D human pose data. We further apply our self-supervised correction mechanism to develop a 3D human pose machine, which jointly integrates the 2D spatial relationship, temporal smoothness of predictions and 3D geometric knowledge. Extensive evaluations on the Human3.6M and HumanEva-I benchmarks demonstrate the superior performance and efficiency of our framework over all the compared competing methods. Please find the code of this
project at: \url{http://www.sysu-hcp.net/3d_pose_ssl/}
\end{abstract}

}

\maketitle

\IEEEdisplaynontitleabstractindextext

%
\IEEEpeerreviewmaketitle

\IEEEraisesectionheading{\section{Introduction}\label{sec:introduction}}

\begin{figure}[t]
\centering
\includegraphics[width=0.99 \columnwidth]{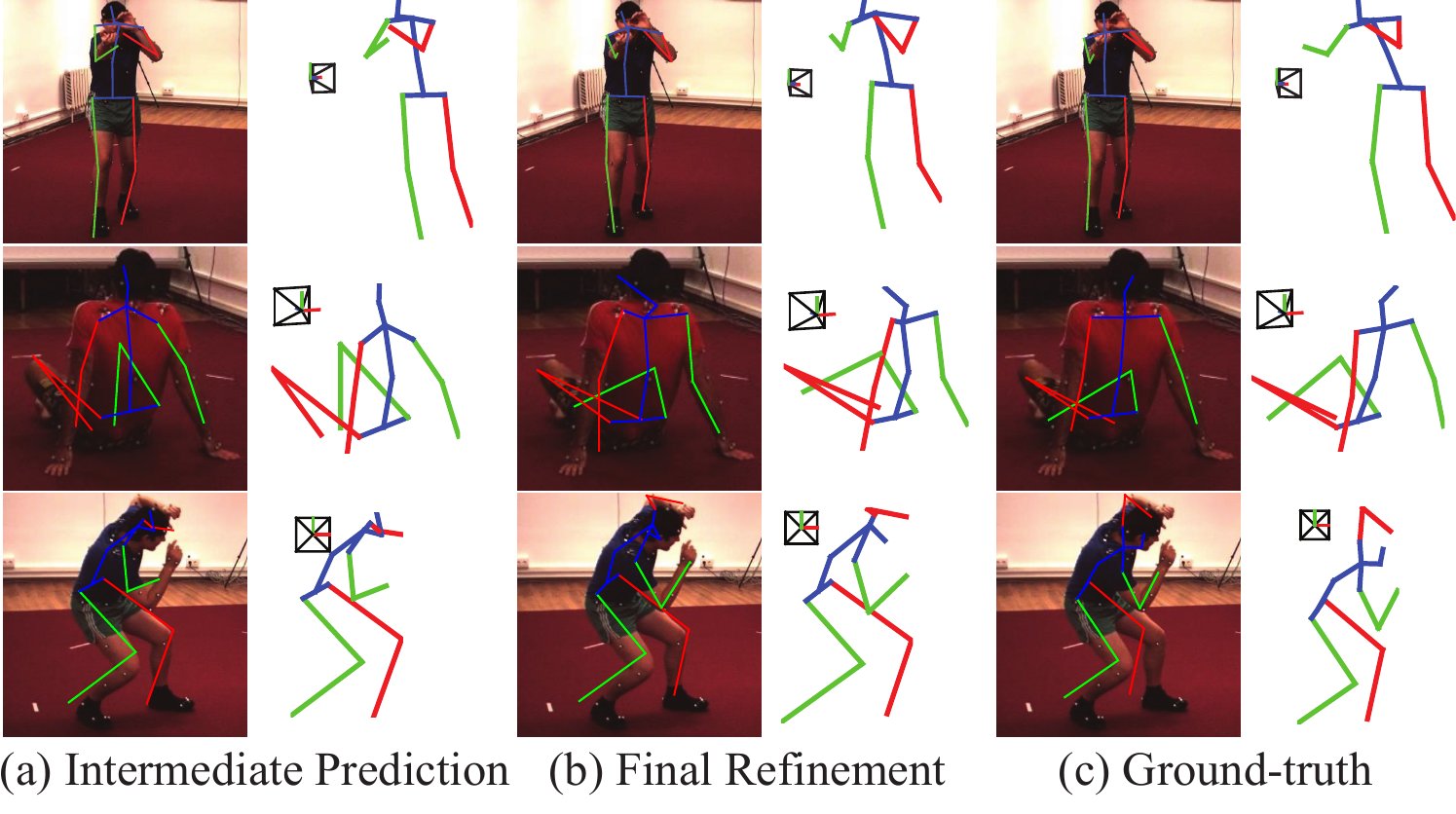}
\caption{Some visual results of our approach on \keze{the} Human3.6M \keze{benchmark}~\cite{huamn3.6m}. (a) illustrates the intermediate 3D poses estimated by the 2D-to-3D pose transformer module, (b) denotes the final 3D poses refined by the 3D-to-2D pose projector module, \kezen{and} (c) denotes the ground-truth. The estimated 3D joints are reprojected into the images and shown by themselves from the side view (next to the images). As shown, \keze{the predicted 3D poses} in (b) \keze{have been} significantly corrected, compared with (a). Best viewed in color. \keze{Note that, red} and green indicate left and right, respectively.}
\label{fig:3d_pose_example}
\end{figure}

%
%
%
%
\IEEEPARstart{R}{ecently}, \keze{estimating 3D full-body human poses from monocular RGB imagery has attracted substantial academic interests for its vast potential on human-centric applications, including human-computer interactions~\cite{errity2016human}, surveillance~\cite{held2012intelligent}, and virtual reality~\cite{zz_rheingold1991virtual}. In fact, estimating human pose from images is quite challenging with respect to large variances in human appearances, arbitrary viewpoints, invisibilities of body parts. Besides, the 3D articulated pose recovery from monocular imagery is considerably more difficult since 3D poses are inherently ambiguous from a geometric perspective~\cite{zhou2016deep}, as shown in Fig.~\ref{fig:3d_pose_example}. }

\keze{Recently, notable successes have been achieved for 2D pose estimation based on 2D part models coupled with 2D deformation priors~\cite{xiaohan2015joint,yang2011articulated}, and the deep learning techniques~\cite{deeppose, deepfacialpoint, wei2016convolutional,yang2016end}.} Driven by these successes,
\keze{some 3D pose estimation works~\cite{andriluka2010monocular,zhou2014spatio,zhou2015sparseness,Tekin_2016_CVPR,pem17CVPR,ds16CVPR} attempt to leverage the state-of-the-art 2D pose network architectures (e.g., Convolutional Pose Machines (CPM)~\cite{wei2016convolutional} and Stacked Hourglass Networks~\cite{newell2016stacked}) by combing the image-based 2D part detectors, 3D geometric pose priors and temporal models. These attempts mainly follow \kezen{three} types of pipelines.}
\keze{The first type~\cite{pavlakos2017volumetric,lfd17CVPR,rpsm17CVPR} focuses on directly recovering 3D human poses from 2D input images by utilizing the state-of-the-art 2D pose network architecture to extract 2D pose-aware features with separate techniques and prior knowledge. In this way, these methods can employ sufficient 2D pose annotations to improve the shared feature representation of the 3D pose and 2D pose estimation tasks. The second type~\cite{pem17CVPR,yasin2016dual,weak17ICCV} concentrates on learning a 2D-to-3D pose mapping function. Specifically, the methods\keze{belonging} to this kind first extract 2D poses from 2D input images and further perform 3D pose reconstruction/regression based on these 2D pose predictions. \kezen{The third type~\cite{bogo2016keep,omran2018nbf,hmrKanazawa17} aims at integrating the Skinned Multi-Person Linear (SMPL) model~\cite{SMPL2015} within a deep network to reconstruct 3D human pose and shape in a full 3D mesh of human bodies.} Although having achieved a promising performance, \kezen{all of} these \kezen{kinds} suffer from the heavy computational cost \kezen{by using the time-consuming network architecture (e.g., ResNet-50~\cite{he2015deep})} and limited scalability for all scenarios due to the insufficient 3D pose data.}


\keze{To address the above-mentioned issues and utilize the sufficient 2D pose data for training}, we propose an effective yet efficient 3D human pose estimation framework, which \keze{implicitly learns to integrate the 2D spatial relationship, temporal coherency and 3D geometry knowledge by utilizing the advantages afforded by Convolutional Neural Networks (CNNs)~\cite{wei2016convolutional} (i.e., the ability to learn feature representations for both image and spatial context directly from data), recurrent neural networks (RNNs)~\cite{Hochreiter1997Long} (i.e., the ability to model the temporal dependency and prediction smoothness) and the self-supervised correction (i.e., the ability to implicitly retain 3D geometric consistency between the 2D projections of 3D poses and the predicted 2D poses). Concretely,} our model employs a sequential training to capture long-range temporal coherency among multiple human body parts, and it is further enhanced via a novel self-supervised correction mechanism, which involves two dual learning tasks, i.e., 2D-to-3D pose transformation and 3D-to-2D pose projection, to generate geometrically consistent 3D pose predictions under a \keze{self-supervised} correction mechanism, i.e., forcing the 2D projections of the generated 3D poses to be identical to the estimated 2D poses.

As illustrated in Fig.~\ref{fig:3d_pose_example}, our model enables \keze{the} gradual refinement of the 3D pose prediction for each frame according to the coherency of sequentially predicted 2D poses and 3D poses, contributing to seamlessly learning the pose-dependent \kezen{constraints} among multiple body parts and sequence-dependent context from the previous frames. Specifically, \keze{taking} each frame as \kezen{input}, \keze{our model} first extracts the 2D pose representations and predicts the 2D poses. Then, the 2D-to-3D pose transformer module is injected to transform the learned pose representations from the 2D domain to the 3D domain, and it further regresses the intermediate 3D poses via two stacked long short-term memory (LSTM) layers by combining the following two lines of information, \ie, the transformed 2D pose representations and the learned states from past frames. Intuitively, the 2D pose representations are conditioned on the monocular image, which captures the spatial appearance and context information. Then, temporal contextual dependency is captured by the hidden states of LSTM units, which effectively improves the robustness of the 3D pose estimations over time. Finally, the 3D joint prediction implicitly encodes the 3D geometric structural information by the 3D-to-2D pose projector module under the introduced \keze{self-supervised} correction mechanism. In specific, considering that the 2D projections of 3D poses and the predicted 2D poses should be identical, the minimization of their dissimilarities is regarded as a learning objective for the 3D-to-2D pose projector module to \keze{bidirectionally} correct (or refine) the intermediate 3D pose predictions. Through this \keze{self-supervised} correction mechanism, our model is capable of effectively achieving geometrically coherent 3D human pose predictions without requesting additional 3D \keze{pose} annotations. Therefore, our introduced correction mechanism is self-supervised, and can enhance our model by adding the external large-scale 2D human pose data into the training \kezen{process} to cost-effectively \kezen{increase} the \keze{3D pose} estimation performance.

The main \textbf{contributions} of this work are three-fold. i) We present a novel model that learns to integrate rich spatial and temporal long-range dependencies as well as 3D geometric constraints, rather than relying on specific manually defined body smoothness or kinematic constraints; ii) Developing a simple yet effective \keze{self-supervised} correction mechanism to incorporate 3D pose geometric structural information is innovative in literature, \kezen{and} may also inspire other 3D vision tasks; iii) The proposed self-supervised correction mechanism enables our model to significantly improve 3D human pose estimation via sufficient 2D human pose data. Extensive evaluations on the public challenging Human3.6M~\cite{huamn3.6m} and HumanEva-I~\cite{sigal2010humaneva} benchmarks \keze{demonstrate the superiority of our framework over all the compared \kezen{competing} methods.}

The remainder of this paper is organized as follows. Section~\ref{sec:related} briefly reviews the existing 3D human pose estimation approaches that motivate this work. Section~\ref{sec:alg} presents the details of the proposed model, with a thorough analysis of every component. Section~\ref{sec:exper} presents the experimental results on two public benchmarks with comprehensive evaluation protocols, as well as comparisons with \kezen{competing} alternatives. Finally, Section~\ref{sec:conclude} concludes this paper.

\section{Related Work}
\label{sec:related}
Considerable research has addressed the challenge of 3D human pose estimation. Early research on 3D monocular pose estimation from videos \kezen{involved} frame-to-frame pose tracking and dynamic models that rely on Markov dependencies among previous frames, \eg, \cite{wang2014robust,sigal2012loose}. The main drawbacks of these approaches are the requirement of the initialization pose and the inability to recover from tracking failure. To overcome these drawbacks, more recent approaches \cite{andriluka2010monocular,BurgosArtizzuBMVC13PoseNms} focus on detecting candidate poses in each individual frame, and a post-processing step attempts to establish temporally consistent poses. Yasin \etal \cite{yasin2016dual} proposed a dual-source approach for 3D pose estimation from a single image. They combined the 3D pose data from a motion capture system with an image source annotated with 2D poses. They transformed the estimation into a 3D pose retrieval problem. One major limitation of this approach is its time efficiency. Processing an image requires more than 20 seconds. Sanzari \etal \cite{DBLP:conf/eccv/SanzariNP16} proposed a hierarchical Bayesian non-parametric model, which relies on a representation of the idiosyncratic motion of human skeleton joint groups, and the consistency of the connected group poses is \kezen{considered} during the pose reconstruction.

Deep learning has recently demonstrated its capabilities in many computer vision tasks, such as 3D human pose estimation. Li and Chan \cite{li20143d} first used CNNs to regress the 3D human pose from monocular images and proposed two training strategies to optimize the network. Li \etal \cite{li2015maximum} proposed integrating the structure learning into a deep learning framework, which consists of a convolutional neural network to extract image features and two following subnetworks to transform the image features and poses into a joint embedding. Tekin \etal \cite{Tekin_2016_CVPR} proposed exploiting motion information from consecutive frames and applied a deep learning network to regress the 3D pose. Zhou \etal \cite{zhou2015sparseness} proposed a 3D pose estimation framework from videos that consists of a novel synthesis among a deep-learning-based 2D part detector, a sparsity-driven 3D reconstruction approach and a 3D temporal smoothness prior. Zhou \etal \cite{zhou2016deep} proposed directly embedding a kinematic object model into the deep learning network. Du \etal \cite{DBLP:conf/eccv/DuWLHGWKG16} introduced additional built-in knowledge for reconstructing the 2D pose and formulated a new objective function to estimate the 3D pose from the detected 2D pose. More recently, Zhou \etal \cite{pavlakos2017volumetric} presented a coarse-to-fine prediction scheme to cast 3D human pose estimation as a 3D keypoint localization problem in a voxel space in an end-to-end manner. Moreno-Noguer \etal \cite{DMR17CVPR} formulated the 3D human pose estimation problem as a regression between matrices encoding 2D and 3D joint distances. Chen \etal \cite{pem17CVPR} proposed a simple approach to 3D human pose estimation by performing 2D pose estimation followed by 3D exemplar matching. Tome \etal \cite{lfd17CVPR} proposed a multi-task framework to jointly integrate 2D joint estimation and 3D pose reconstruction to improve both tasks. To leverage the well-annotated large-scale 2D pose datasets, Zhou \etal \cite{weak17ICCV} proposed a weakly-supervised transfer learning method that uses mixed 2D and 3D labels in a unified deep \keze{two-stage cascaded structure network}. However, these methods oversimplify the 3D geometric knowledge. In contrast to all these \kezen{aforementioned} methods, our model can leverage a lightweight network architecture to implicitly learn to integrate the 2D spatial relationship, temporal coherency and 3D geometry knowledge in a fully differential manner.

\kezen{Instead of directly computing 2D and 3D joint locations, several works concentrate on producing a 3D mesh body representation by using a CNN to predict Skinned Multi-Person Linear model \cite{SMPL2015}. For instance, Omran \etal \cite{omran2018nbf} proposed to integrate a statistical body model within a CNN, leveraging reliable bottom-up semantic body part segmentation and robust top-down body model constraints. Kanazawa \etal \cite{hmrKanazawa17} presented an end-to-end adversarial learning framework for recovering a full 3D mesh model of a human body by parameterizing the mesh in terms of 3D joint angles and a low dimensional linear shape space. Furthermore, this method employs the weak-perspective camera model to project the 3D joints onto the annotated 2D joints via an iterative error feedback loop~\cite{feedback16}. Similar to our proposed method, these approaches also regard the in-the-wild images with 2D ground-truth as the supervision to improve the model performance. The main difference is that our self-supervised learning method is more flexible and robust without relying on the assumption of the weak-perspective camera model.}

\begin{figure*}[t]
\centering
  \includegraphics[width=0.85\linewidth]{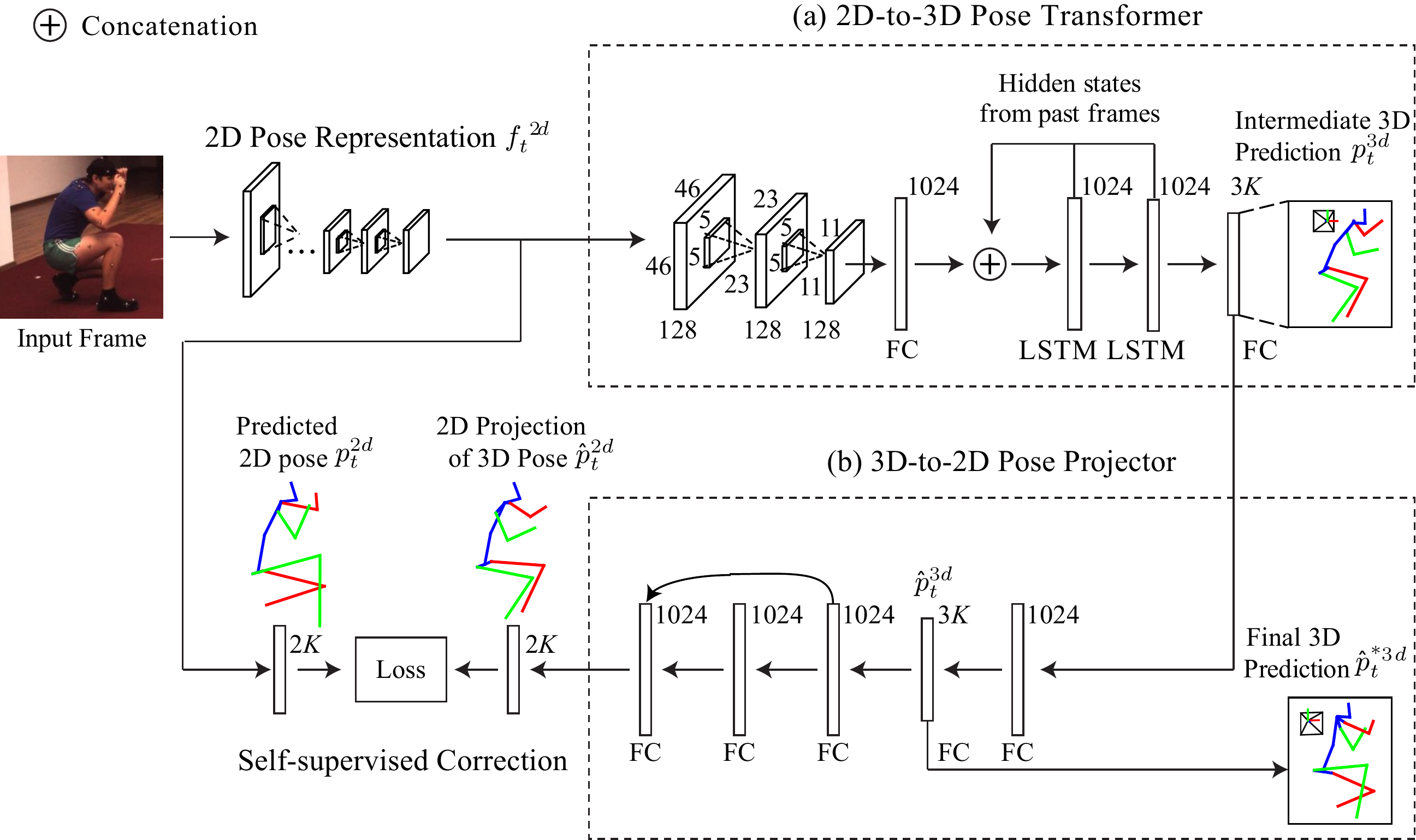}
  \vspace{-5pt}
  \caption{An overview of the proposed 3D human pose machine framework. Our model predicts the 3D human poses for the \keze{given} monocular image frames, and it \keze{progressively} refines its predictions with \keze{the proposed} self-supervised correction. Specifically, the estimated 2D pose \keze{$p_t^{2d}$} with the \keze{corresponding} pose representation \keze{$f_t^{2d}$} for \keze{each} frame of the input sequence \keze{is} first obtained and further passed into two neural network modules: i) a 2D-to-3D pose transformer module for transforming the pose representations from the 2D domain to the 3D domain to intermediately predict the human joints \keze{$p_t^{3d}$} in the 3D coordinates, and ii) a 3D-to-2D pose projector module \keze{to obtain the projected 2D pose \keze{$\hat{p}_t^{2d}$ after regressing $p_t^{3d}$ into $\hat{p}_t^{3d}$}. Through minimizing the difference between $p_t^{2d}$ and $\hat{p}_t^{2d}$, our model is capable of bidirectionally refining the regressed 3D poses \keze{$\hat{p}_t^{3d}$} via the proposed self-supervised correction mechanism}. Note that the parameters of the 2D-to-3D pose transformer module for all frames are shared to preserve the temporal motion coherence. \keze{$3K$ and $2K$ denotes the dimension of the vector for representing the 3D and 2D human pose formed by $K$ skeleton joints, respectively.}}\label{fig:framework_detail}
  \vspace{-12pt}
\end{figure*}

\keze{Our approach is close to\cite{lfd17CVPR}, which also used the projection from the 3D space to the 2D space to improve the 3D pose estimation performance. However, there are two main differences between \cite{lfd17CVPR} and our model: i) The definition of the 3D-to-2D projection function and the optimization strategy. Rather than explicitly defining a concrete model, our 3D-to-2D projection is implicitly learned in a completely data-driven manner. However, the projection of 3D poses in \cite{lfd17CVPR} is explicitly modeled by using a weak perspective model, which consists of the orthographic projection matrix, a known external camera calibration matrix and an unknown rotation matrix. As claimed in \cite{lfd17CVPR}, this explicit model is prone to sticking in local minima during the training. Thus, the authors have to quantize over the space of possible rotations. Through this approximation, their model performance may suffer from the fixed choices of rotations; ii) The way of utilizing the projected 2D pose. In contrast to \cite{lfd17CVPR} which learns to weightily fuse the projected 2D and the estimated 2D poses for further regressing the final 3D pose, our model exploits the 3D geometric consistency between the projected 2D and the estimated 2D poses to bidirectionally refine the intermediate 3D pose predictions.
}

\keze{
  \textbf{Self-supervised Learning.} Aiming at training the feature representation without relying on manual data annotation, self-supervised learning (SSL) has first been introduced in \cite{Pal_computerrecognition} for vowel class recognition, and further extended for object extraction in ~\cite{Ghosh93self-organizationfor}. Recently, plenty of SSL methods (e.g.,~\cite{MT_2017_ICCV,xl_2015_ICCV}) have been proposed. For instance, \cite{MT_2017_ICCV} investigated multiple self-supervised methods to encourage the network to factorize the information in its representation. In contrast to these methods that focus on learning an optimal visual representation, our work considers the self-supervision as an optimization guidance for 3D pose estimation.
}

Note that a preliminary version of this work \kezen{was} published in \cite{rpsm17CVPR}, which uses multiple stages to gradually refine the predicted 3D poses. The network parameters \kezen{in the} multiple stages are recurrently trained in a fully end-to-end manner. However, the multi-stage mechanism results in a heavy computational cost, and the stage-by-stage improvement \kezen{is} less significant as the number of stages increases. In this paper, we inherit its idea of integrating the 2D spatial relationship, temporal coherency as well as 3D geometry knowledge, and we further impose a novel self-supervised correction mechanism to further enhance our model by bridging the domain gap between the 3D and 2D human poses. Specifically, we develop a 3D-to-2D pose projector module to replace the multi-stage refinement to correct the intermediate 3D pose predictions by retaining the 3D geometric consistency between their 2D projections and the predicted 2D poses. Therefore, the imposed correction mechanism enables us to leverage the external large-scale 2D human pose data to boost 3D human pose estimation. Moreover, more comparisons with \kezen{competing} approaches and more detailed analyses of the proposed modules are included to further verify our statements.

\section{3D Human Pose Machine}
\label{sec:alg}
We propose a 3D human pose machine to resolve 3D pose sequence generation for monocular frames, and \kezen{we} introduce a concise \keze{self-supervised correction} mechanism to enhance our model by retaining the 3D geometric consistency. After extracting \kezen{the} 2D pose representation and estimating \kezen{the} 2D poses for each frame via a common 2D pose sub-network, our model employs two consecutive modules. \kezen{The first module is} {\em the 2D-to-3D pose transformer module} \kezen{for transforming} the 2D pose-aware features from the 2D domain to the 3D domain. This module is designed to estimate intermediate 3D poses for each frame by incorporating temporal dependency in the image sequence. \kezen{The second module} is {\em the 3D-to-2D pose projector module} \kezen{for} \keze{bidirectionally} \kezen{refining} the intermediate 3D pose prediction via our introduced \keze{self-supervised} correction mechanism. These two modules are combined in a unified framework to be optimized in a fully end-to-end manner.

\begin{figure*}[t]
  \centering
  \includegraphics[width=1 \textwidth]{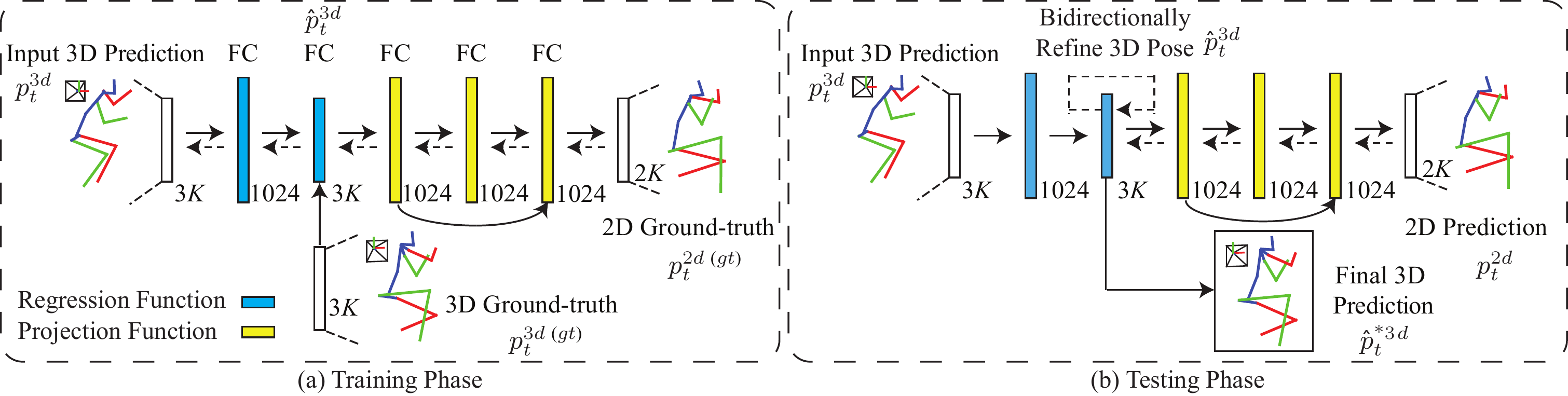}
  \vspace{-15pt}
  \caption{Detailed sub-network architecture of our proposed 3D-to-2D pose projector module in the (a) training phase and (b) testing phase. The Fully Connected (FC) layers for the regression function are in blue, while those for the projection function are in yellow. The black arrows represent the forward data flow, while the dashed arrows denote the backward propagation \kezen{used} to update the network parameters and perform \keze{gradual} pose refinement in (a) and (b), respectively.}
  \label{fig:3dto2d_detail}
  \vspace{-10pt}
\end{figure*}

As illustrated in Fig.~\ref{fig:framework_detail}, our model performs the sequential refinement with self-supervised correction to generate the 3D pose sequence. Specifically, the $t$-th frame $I_t$ is passed into the 2D pose sub-network $\Psi_R$, the 2D-to-3D pose transformer module $\Psi_T$, and the 3D-to-2D projector module \keze{$\{\Psi_C, \Psi_P\}$} to predict the final 3D poses. The 2D pose sub-network is stacked by convolutional and fully connected layers, \kezen{and} the 2D-to-3D pose transformer module contains two LSTM layers to capture the temporal dependency over frames. \keze{Specifically}, given the input image sequence with $N$ frames, \kezen{the} 2D pose sub-network $\Psi_R$ is first employed to extract the 2D pose-aware features $f_{t}^{2d}$ and predict the 2D pose $p_{t}^{2d}$ for the $t$-th frame of the input sequence. Then, the extracted 2D pose-aware features $f_{t}^{2d}$ are further fed into the 2D-to-3D pose transformer module $\Psi_T$ to obtain the intermediate 3D pose $p_{t}^{3d}$ \keze{, where $\Psi_T$} is composed of the hidden states $H_{t-1}$ learned from the past frames. Finally, the predicted 2D poses $p_{t}^{2d}$ and intermediate 3D pose $p_{t}^{3d}$ are fed into the 3D-to-2D projector module with two functions, i.e., $\Psi_C$ and $\Psi_P$, to obtain the final 3D poses \keze{$\hat{p}_{t}^{*3d}$}. \kezen{Considering that} most existing 2D human pose data are \kezen{still images without temporal orders}, \keze{we additionally introduce a simple yet effective regression function $\Psi_C$ to transform the intermediate 3D pose vector $p_{t}^{3d}$ into a changeable prediction $\hat{p}_{t}^{3d}$.} The projection function $\Psi_P$ implies projecting the 3D coordinate $\hat{p}_{t}^{3d}$ into the image plane to obtain the projected 2D pose $\hat{p}_{t}^{2d}$. Formally, $f^{2d}_{t}$, $p_{t}^{3d}$, $\hat{p}_{t}^{3d}$, and $\hat{p}_{t}^{2d}$ are formulated as follows:
\begin{equation}
\begin{aligned}
\label{eq:opt}
\{ f^{2d}_{t}, p_{t}^{2d} \} &= \Psi_R(I_t; \omega_R), \\
p_{t}^{3d} &= \Psi_T(f^{2d}_{t}; \omega_T, H_{t-1}), \\
\hat{p}_{t}^{3d} &= \Psi_C(p_{t}^{3d}; \omega_C), \\
\hat{p}_{t}^{2d} &= \Psi_P(\hat{p}_{t}^{3d}; \omega_P), \\
\end{aligned}
\end{equation}
where $\omega_R$, $\omega_T$, $\omega_C$ and $\omega_P$ are parameters of $\Psi_R, \Psi_T, \Psi_C$ and $\Psi_P$, respectively. Note that, $H_0$ is initially set to be a vector of zeros. After obtaining the predicted 2D pose $p_{t}^{2d}$ via $\Psi_R$, and the projected 2D pose $\hat{p}_{t}^{2d}$ via $\Psi_P$ in Eq.~(\ref{eq:opt}), we consider minimizing the dissimilarity between $p_{t}^{2d}$ and $\hat{p}_{t}^{2d}$ as an optimization objective to obtain the optimal $\hat{p}_{t}^{*3d}$ for the $t$-th frame.

\begin{table*}[t]
\center
\caption{Details of the convolutional layers in the 2D pose sub-network.}
\label{table:shared_network_details}
\vspace{-5pt}
\begin{adjustbox}{max width=1.0\textwidth}
\begin{tabular}{|c|c|c|c|c|c|c|c|c|c|}
\hline
 & 1 & 2 & 3 & 4 & 5 & 6 & 7 & 8 & 9 \\ \hline
Layer Name & conv1\_1 & conv1\_2 & max\_1 & conv2\_1 & conv2\_2 & max\_2 & conv3\_1 & conv3\_2 & conv3\_3 \\ \hline
Channel (kernel-stride) & 64(3-1) & 64(3-1) & 64(2-2) & 128(3-1) & 128(3-1) & 128(2-2) & 256(3-1) & 256(3-1) & 256(3-1) \\ \hline
 & 10 & 11 & 12 & 13 & 14 & 15 & 16 & 17 & 18 \\ \hline
Layer Name & conv3\_4 & max\_3 & conv4\_1 & conv4\_2 & conv4\_3 & conv4\_4 & conv4\_5 & conv4\_6 & conv4\_7 \\ \hline
Channel (kernel-stride) & 256(3-1) & 256(2-2) & 512(3-1) & 512(3-1) & 256(3-1) & 256(3-1) & 256(3-1) & 256(3-1) & 128(3-1) \\ \hline
\end{tabular}
\end{adjustbox}
\vspace{-10pt}
\end{table*}

In the following, we will introduce more details of our model and \kezen{provide} comprehensive clarifications to make \kezen{the work easier} to understand. The corresponding algorithm for jointly training these modules will also be discussed at the end.

\subsection{2D Pose Sub-network}
\label{sec:2d_pose_cnn}
The objective of the 2D pose sub-network is to encode each frame in \keze{a given} monocular sequence with a compact representation of the pose information, \eg, the body shape of the human. \kezen{The} shallow convolution layers often extract the common low-level information, which is a very basic representation of the human image. \keze{We build our 2D pose sub-network by borrowing the architecture of the convolutional pose machines~\cite{wei2016convolutional}}. \keze{Please see Table~\ref{table:shared_network_details} for more details. Note that other state-of-the-art architectures for 2D pose estimation can be also utilized. As illustrated in Fig.~\ref{fig:framework_detail}, the 2D pose sub-network} takes the $368\times 368$ image as input, and it outputs the 2D pose-aware feature maps with a size of $128\times 46\times 46$ and the predicted 2D pose vectors \kezen{with 2$K$ entries being the argmax positions of these feature maps}.

\subsection{2D-to-3D Pose Transformer Module}
Based on the features extracted by the 2D pose sub-network, the 3D pose transformer module is employed to adapt the 2D pose-aware features in an adapted feature space for the later 3D pose prediction. As depicted in Fig.~\ref{fig:framework_detail} (a), two convolutional layers and one fully connected layer are leveraged. Each convolutional layer contains 128 different kernels with a size of $5\times5$ and a stride of 2, and a max pooling layer with a $2\times2$ kernel size and a stride of 2 is appended on the convolutional layers. Finally, the convolution features are fed to a fully connected layer with 1024 units to produce the adapted feature vector. In this way, the 2D pose-aware features are transformed into the 1024-dimensional adapted feature vector.


Given the adapted features for all frames, we employ LSTM to sequentially predict the 3D pose sequence by incorporating rich temporal motion patterns among frames as \cite{rpsm17CVPR}. Note that, LSTM \cite{Hochreiter1997Long} has been proven to \kezen{achieve} better performance \kezen{in} exploiting temporal correlations than a vanilla recurrent neural network in many tasks, \eg, speech recognition \cite{graves2014towards} and video description \cite{donahue2015long}. In our model, we \kezen{use} the LSTM layers to capture the temporal dependency in the monocular sequence for refining the 3D pose prediction for each frame. As illustrated in Fig.~\ref{fig:framework_detail} (b), our model employs two LSTM layers with 1024 hidden cells and an output layer that predicts the locations of $K$ joint points of the human. In particular, the hidden states learned by the LSTM layers are capable of implicitly encoding the temporal dependency across different frames of the input sequence. As formulated in Eq.~(\ref{eq:opt}), incorporating the previous hidden states imparts our model with the ability to sequentially refine the pose predictions.

\subsection{3D-to-2D Projector Module}
As illustrated in Fig.~\ref{fig:3dto2d_detail} (a), this module consists of six fully connected (FC) layers \kezen{containing} ReLU and batch normalization operations. As one can see from left to right in Fig.~\ref{fig:3dto2d_detail}(a), the first two FC layers (denoted in blue)
\kezen{define} the regression function $\Psi_C$ in which the intermediate 3D pose predictions are regressed into the pose prediction $\hat{p}_{t}^{3d}$, and the remaining four FC layers (denoted in yellow) with 1024 units represent the projection function $\Psi_P$ that projects $\hat{p}_{t}^{3d}$ into the image plane to obtain the projected 2D pose $\hat{p}_{t}^{2d}$. Moreover, an identical mapping as ResNet~\cite{he2015deep} is used inside $\Psi_P$ to make the information pass through quickly to avoid overfitting. Therefore, our 3D-to-2D projector module is simple yet powerful for both regression and projection tasks. \kezen{Considering the self-corrected 3D pose may need to be discarded sometimes, we regard the regression function $\Psi_C$ as a copy to be corrected for the intermediate 3D poses.}

In the training phase, \keze{we first initialize the module parameters \{$\omega_C$, $\omega_P$\} for $\Psi_C$ and $\Psi_P$ via the supervision of the 3D and 2D ground-truth poses from 3D human pose data as illustrated in Fig.~\ref{fig:3dto2d_detail} (a), respectively. The optimization function is:
\begin{small}\begin{equation}
\label{eq:loss_proj}
\begin{gathered}
 \min_{\{\omega_C, \omega_P\}} \sum_{t=1}^{N} \left \| \hat{p}_t^{3d} - p_t^{3d(gt)} \right \|_{2}^{2} +
\left \| \Psi_P(\hat{p}_t^{3d}; \omega_P) - p_{t}^{2d(gt)} \right \|_{2}^{2}, \\
 \end{gathered}
\end{equation}\end{small}where $\hat{p}_t^{3d}$ is the regressed 3D pose via $\Psi_C$ in Eq.~(\ref{eq:opt}), and its 2D projection is $\hat{p}_t^{2d}=\Psi_P(\hat{p}_t^{3d}; \omega_P)$. Eq.~(\ref{eq:loss_proj}) forces $\Psi_C$ to regress $\hat{p}_t^{3d}$ from intermediate 3D poses $p_t^{3d}$ to the 3D pose ground-truth $p_t^{3d(gt)}$, and it further forces the output of $\Psi_P$, i.e., the projected 2D poses, to be similar to the 2D pose ground-truth $p_{t}^{2d(gt)}$. In this way, the 3D-to-2D pose projector module can learn the geometric consistency to correct intermediate 3D pose predictions.} After initialization, we substitute the predicted 2D poses and 3D poses for \kezen{the} 2D and 3D ground-truth to optimize $\Psi_C$ and $\Psi_P$ in a self-supervised fashion. Considering that the predictions for \kezen{certain} body joints (e.g., hand$_{left}$, hand$_{right}$, foot$_{left}$ and foot$_{right}$ defined in the Human3.6M dataset) may not be accurate and reliable due to the challenging \kezen{nature} of the rich flexibilities and occlusions of body joints, we employ the dropout trick~\cite{krizhevsky2012imagenet} \kezen{in} the intermediate 3D pose estimations $p_{t}^{3d}$ and the predicted 2D pose $p_{t}^{2d}$, i.e., the position for each body joint has a probability $\delta$ to be zero. This trick enables the regression function $\Psi_C$ and the project function $\Psi_P$ to be insensitive to the outliers inside $p_{t}^{3d}$ and $p_{t}^{2d}$. \kezen{As reported in~\cite{krizhevsky2012imagenet}, the dropout trick can significantly contribute to alleviating the overfitting of the fully connected layers inside $\Psi_C$ and $\Psi_P$.}
Meanwhile, we also \kezen{employ} the 3D pose ground-truth to encourage the regression function $\Psi_C$ to learn to regress the  3D pose estimation $\hat{p}_{t}^{3d}$. In our experiments, $\delta$ is empirically set to be 0.3.

\keze{
The inference phase of this module is also self-supervised. Specifically, given the predicted 2D pose $p_{t}^{2d}$, we can obtain the initial $\hat{p}_{t}^{3d}$ and the corresponding projected 2D pose $\hat{p}_{t}^{2d}$ via forward propagation, as indicated in Fig.~\ref{fig:3dto2d_detail} (b). According to the 3D geometric consistency that the projected 2D pose $\hat{p}_{t}^{2d}$ should be identical to the predicted 2D pose $p_{t}^{2d}$, we propose minimizing the dissimilarity between $\hat{p}_{t}^{2d}$ and $p_{t}^{2d}$ by optimizing its specific $\omega_P^t$ and $\omega_C^t$ as follows:
\kezen{\begin{equation}
\label{eq:inference}
\begin{split}
\{\omega_{P}^{*t}, \omega_{C}^{*t}\} &= \argmin_{\{\hat{p}_{t}^{3d}, \omega_P^t\}} \| p_{t}^{2d} - \hat{p}_{t}^{2d} \|_2^2 \\
 &= \argmin_{\{\hat{p}_{t}^{3d}, \omega_P^t\}}  \| p_{t}^{2d} - \Psi_P( \hat{p}_{t}^{3d}; \omega_P^t) \|_{2}^{2} \\
 &= \argmin_{\{\omega_{P}^t, \omega_C^t\}}  \| p_{t}^{2d} - \Psi_P( \Psi_C(p_{t}^{3d}; \omega_C^t); \omega_P^t) \|_{2}^{2}, \\
 \end{split}
\end{equation}}where the parameters \{$\omega_P^{t}$, $\omega_C^{t}$\} are initialized from the \keze{well-optimized} \{$\omega_P$, $\omega_C$\} from the training phase. \keze{Note that, $\omega_C^{*t}$ and $\omega_P^{*t}$ are disposable and only valid for $I_t$.} Since $p_{t}^{2d}$ and $p_{t}^{3d}$ are fixed, we first perform forward propagation to obtain the initial prediction, and further employ the standard back-propagation algorithm~\cite{CNN1990} to obtain $\omega_C^{*t}$ and $\omega_P^{*t}$ via Eq.~(\ref{eq:inference}). Thus, the output 3D pose regression $\hat{p}_{t}^{3d}$ is bidirectionally refined to be the final 3D pose prediction during the the optimizing of $\omega_C^{t}$ and $\omega_P^{t}$ according to the proposed self-supervised correction mechanism.} At the end, the final 3D pose $\hat{p}_{t}^{*3d}$ is obtained according to Eq.~(\ref{eq:inference}) as follows:
\begin{equation}
\label{eq:final}
\hat{p}_{t}^{*3d} = \Psi_C(p_{t}^{3d}; \omega_C^{*t}).
\end{equation}

The hyperparameters (i.e., the iteration number and learning rate) for \kezen{$\omega_P^{*t}$} and \kezen{$\omega_C^{*t}$} play a crucial role in effectively and efficiently refining the 3D pose estimation. In fact, a large iteration number and small learning rate can ensure that the model is capable of converging to a satisfactory $\hat{p}_{t}^{*3d}$. However, this setting results in a heavy computational cost. Therefore, a small iteration number with large learning rate is preferred to achieve a trade-off between efficiency and accuracy. Moreover, although we can achieve high accuracy on 2D pose estimation, the predicted 2D poses may contain errors due to the heavy occlusion of human body parts. Treating these inaccurate 2D poses as optimization objectives to \keze{bidirectionally} refine the 3D pose prediction \keze{is prone to a decrease} in performance. To address this issue, we utilize a heuristic strategy to determine the optimal hyperparameters used for each frame in our implementation. Specifically, we can check the convergence of some robust skeleton joints (i.e., Pelvis, Shoulder$_{left}$, Shoulder$_{right}$, Hip$_{left}$ and Hip$_{right}$ defined in the Human3.6M dataset) in each iteration. In practice, we find that the predictions of these reliable joints are generally less flexible and have \kezen{lower} probabilities of being occluded than other joints. If the predicted 2D pose contains small errors, then these joints of the refined 3D pose $\hat{p}_{t}^{*3d}$ will have large \kezen{and} inconsistent changes \kezen{within} the self-supervised correction. Hence, we terminate the further refinement when the positions of these joints are converged (i.e., average changes $< \epsilon$mm), and discard the self-supervised correction when the average change \kezen{in} these joints are not within an empirical threshold $\tau$mm. In our experiments, we empirically set $\{\tau, \epsilon\}=\{20, 5\}$ and employ two back-propagation \kezen{operations} to update $\omega_P$ and $\omega_C$ before outputting the final 3D pose prediction $\hat{p}_{t}^{*3d}$.

\subsection{Model Training}
In the training phase, the optimization of our proposed model occurs in a fully end-to-end manner, and we have defined several types of loss functions to fine-tune the network parameters: $\omega_R, \omega_T$ and \{$\omega_C, \omega_P$\}, respectively. For the 2D pose sub-network, we build an extra FC layer upon the convolutional layers of the 2D pose sub-network to generate 2$K$ joint location coordinates. We leverage the Euclidean distances between the \kezen{predictions} for all $K$ \keze{body joints} and the corresponding ground-truth to train $\omega_R$. Formally, we have:
\begin{equation}
\label{eq:loss_2d}
\min_{\omega_R}  \sum_{t=1}^{N} \left \| p_t^{2d}(gt) - \Psi_R(I_t; \omega_R) \right \|_2^2,
\end{equation}
where $p_t^{2d}(gt)$ denotes the 2D pose ground-truth for the $t$-th frame $I_t$.

For the 2D-to-3D pose transformer module, our model enforces the 3D pose sequence prediction loss for all frames, which is also defined as follows:
\begin{equation}
\label{eq:loss_trans}
\begin{split}
\min_{\omega_T} & \sum_{t=1}^{N} \left \| p_t^{3d} - p_t^{3d(gt)} \right \|_{2}^{2} \\
=& \sum_{t=1}^{N} \left \| \Psi_T( f_t^{2d}; \omega_T, H_{t-1} ) - p_t^{3d(gt)} \right \|_{2}^{2},
\end{split}
\end{equation}
where $p_t^{3d(gt)}$ is the 3D pose ground-truth for the $t$-th frame $I_t$. According to Eq.~(\ref{eq:loss_trans}), we integrally fine-tune the parameters of the 2D-to-3D pose transformer module and the convolutional layers of the 2D pose sub-network in an end-to-end optimization manner. Note that, to obtain sufficient samples to train the 3D pose transformer module, we propose decomposing one long monocular image sequence into several small equal clips with $N$ frames. In our experiments, we jointly feed our model with 2D and 3D pose data after all the network parameters \kezen{are} well initialized. \kezen{For} the 2D human pose data, this module regards their 3D pose sequence prediction loss as zero.


After \keze{initializing the 3D-to-2D projector module via Eq.~(\ref{eq:loss_proj})}, we fine-tune the whole network to jointly optimize the network parameters $\{\omega_R, \omega_T, \omega_C, \omega_P\}$ in a fully end-to-end manner as follows:
\begin{equation}
\label{eq:whole}
\begin{split}
& \min_{\{\omega_R, \omega_T, \omega_C, \omega_P\}} \| p_t^{2d} - \hat{p}_t^{2d} \|_2^2. \\
\end{split}
\end{equation}

\begin{algorithm}[t]
\caption{The Proposed Training Algorithm} \label{alg:alg_overview}
\begin{algorithmic}[1]
\REQUIRE 3D human pose data $\{\mathbf{I}_{t}^{3d}\}_{t=1}^{N}$ and 2D human pose data $\{\mathbf{I}_{i}^{2d}\}_{i=1}^{M}$
\STATE Pre-train the 2D pose sub-network with $\{\mathbf{I}_{i}^{2d}\}_{i=1}^{M}$ to initialize $\omega_R$ via Eq.~(\ref{eq:loss_2d}); \\
\STATE Fixing $\omega_R$, initialize $\omega_T$ with hidden variables $H$ with $\{\mathbf{I}_{t}^{3d}\}_{t=1}^{N}$ via Eq.~(\ref{eq:loss_trans});\\
\STATE Fixing $\omega_R$ and $\omega_T$, initialize \{$\omega_C$, $\omega_P$\} with $\{\mathbf{I}_{t}^{3d}\}_{t=1}^{N}$ via Eq.~(\ref{eq:loss_proj});\\
\STATE Fine-tune the whole model to further update \{$\omega_R$, $\omega_T$, $\omega_C$, $\omega_P$\} on $\{\mathbf{I}_{t}^{3d}\}_{t=1}^{N}$ and $\{\mathbf{I}_{i}^{2d}\}_{i=1}^{M}$ via Eq.~(\ref{eq:whole}).
\RETURN \{$\omega_R$, $\omega_T$, $\omega_C$, $\omega_P$\}.
\end{algorithmic}
\end{algorithm}

Since our model consists of two cascaded modules, the training phase can be divided into the following steps: (i) Initialize the 2D pose representation via pre-training. To obtain \kezen{a} satisfactory feature representation, the 2D pose sub-network is first pre-trained with the MPII Human Pose dataset \cite{andriluka14cvpr}, which \kezen{includes} a larger variety of 2D pose data. (ii) Initialize the 2D-to-3D pose transformer module. We fix the parameters of the 2D pose sub-network and optimize the network parameter $\omega_T$. (iii) Initialize the 3D-to-2D pose projector module. We fix the above optimized parameters and optimize the network parameter \{$\omega_C$, $\omega_P$\}. (iv) Fine-tune the whole model jointly to further update the network parameters \{$\omega_R$, $\omega_T$, $\omega_C$, $\omega_P$\} with the 2D pose and 3D pose training data. For each \kezen{of the} above-mentioned \kezen{steps}, the ADAM~\cite{Kingma2014Adam} strategy is employed for parameter optimization. The entire algorithm can then be summarized as Algorithm \ref{alg:alg_overview}. \kezen{Obviously,} this algorithm is in a good agreement with the pipeline of our model.

\subsection{Model Inference}
In the testing phase, every frame of the input image sequence is sequentially processed via Eq.~(\ref{eq:opt}). Note that each frame $I_t$ has its \keze{own} \{$\omega_C^t$, $\omega_P^t$\} in the 3D-to-2D projector module.  \{$\omega_C^t$, $\omega_P^t$\} are initialized from the well trained \{$\omega_C$, $\omega_P$\}, and \kezen{they} will be updated by minimizing the difference between the predicted 2D poses $p_{t}^{2d}$ and projected 2D poses $\Psi_P( \hat{p}_{t}^{3d}; \omega_P)$ via Eq.~(\ref{eq:inference}). During the inference, the 3D pose estimation is \keze{bidirectionally} refined until convergence is \kezen{achieved} according to the hyperparameter \kezen{settings}. Finally, we output the final 3D pose estimation via Eq.~(\ref{eq:final}).

\begin{table*}[tp]
  \caption{\textbf{Quantitative comparisons on the Human3.6M dataset} using 3D pose errors (in millimeters). The entries with the smallest 3D pose errors for each category are bold-faced. \kezen{A} method with ``*'' denotes that it is individually trained on each action category. Our model achieves a significant improvement over all compared approaches.}\label{table:h3m_res}
  \vspace{-5pt}
\small
\setlength{\tabcolsep}{1pt}
  \begin{center}
  \begin{tabular}{c}
  Protocol \#1 \\
  \end{tabular}
  \end{center}
\begin{adjustbox}{max width=1.0\textwidth}
  \centering
  \begin{tabular}{@{}l|ccccccccccccccc|c@{}}
  \toprule
  \hline
   Method & Direction & Discuss & Eat & Greet & Phone & Pose & Purchase & Sit & SitDown & Smoke & Photo & Wait & Walk & WalkDog & WalkPair & Avg. \\ \midrule
  \hline
  Ionescu PAMI'14\cite{huamn3.6m}*& 132.71 & 183.55 & 132.37 & 164.39 & 162.12 & 150.61 & 171.31 & 151.57 & 243.03 & 162.14 & 205.94 & 170.69 & 96.60 & 177.13 & 127.88 & 162.14 \\
  Li ICCV'15\cite{li2015maximum}* & - & 136.88 & 96.94 & 124.74 & - & - & - & - & - & - & 168.68 &  - & 69.97 & 132.17 & - & - \\
  Tekin CVPR'16 \cite{Tekin_2016_CVPR}* & 102.41 & 147.72 & 88.83 & 125.28 & 118.02 & 112.38 & 129.17 & 138.89 & 224.9 & 118.42 & 182.73 & 138.75 & 55.07 & 126.29 & 65.76 & 124.97 \\
  Zhou CVPR'16 \cite{zhou2015sparseness}* & 87.36 & 109.31 & 87.05 & 103.16 & 116.18 & 106.88 & 99.78 & 124.52 & 199.23 & 107.42 & 143.32 & 118.09 & 79.39 & 114.23 & 97.70 & 113.01 \\
  Zhou ECCVW'16 \cite{zhou2016deep}* & 91.83 & 102.41 & 96.95 & 98.75 & 113.35 & 90.04 & 93.84 & 132.16 & 158.97 & 106.91 & 125.22 & 94.41 & 79.02 & 126.04 & 98.96 & 107.26 \\
  Du ECCV'16 \cite{DBLP:conf/eccv/DuWLHGWKG16}* & 85.07 & 112.68 & 104.90 & 122.05 & 139.08 & 105.93 & 166.16 & 117.49 & 226.94 & 120.02 & 135.91 & 117.65 & 99.26 & 137.36 & 106.54 & 126.47 \\
  Chen CVPR'17 \cite{pem17CVPR}* & 89.87 & 97.57 & 89.98 & 107.87 & 107.31 & 93.56 & 136.09 & 133.14 & 240.12 & 106.65 & 139.17 & 106.21 & 87.03 & 114.05 & 90.55 & 114.18 \\
  Tekin ICCV'17 \cite{Tekin_2017_ICCV}* & 54.23 & 61.41 & 60.17 & 61.23 & 79.41 & 63.14 & 81.63 & 70.14 & \bf 107.31 & 69.29 &\bf 78.31 & 70.27 & 51.79 &74.28 & 63.24 & 69.73 \\
  \textbf{Ours}* &\bf 50.36 & \bf59.74  & \bf54.86  & \bf57.12  & \bf66.30  & \bf53.24  & \bf54.73  & \bf84.58  & 118.49& \bf 63.10 & 78.61 & \bf59.47& \bf 41.96& \bf  64.88 & \bf49.48& \bf 63.79 \\
  \hline
  Sanzari ECCV'16\cite{DBLP:conf/eccv/SanzariNP16} & \textbf{48.82} & \textbf{56.31} & 95.98 & 84.78 & 96.47 & 66.30 & 107.41 & 116.89 & 129.63 & 97.84 & 105.58 & 65.94 & 92.58 & 130.46 & 102.21 & 93.15 \\
  Tome CVPR'17 \cite{lfd17CVPR} & 64.98 & 73.47 & 76.82 & 86.43 & 86.28 & 68.93 & 74.79 & 110.19 & 173.91 & 84.95 & 110.67 & 85.78 & 86.26 & 71.36 & 73.14 & 88.39\\

  Moreno-Noguer CVPR'17\cite{DMR17CVPR} & 69.54 & 80.15 & 78.20 & 87.01 & 100.75 & 76.01 & 69.65 & 104.71 & 113.91 & 89.68 & 102.71 & 98.49 & 79.18 & 82.40 & 77.17 & 87.30 \\
  Lin CVPR'17\cite{rpsm17CVPR} & 58.02 & 68.16 & 63.25 & 65.77 & 75.26 & 61.16 & 65.71 & 98.65 & 127.68 & 70.37 & 93.05 & 68.17 & 50.63 & 72.94 & 57.74 & 73.10 \\
  Pavlakos CVPR'17\cite{pavlakos2017volumetric} & 67.38 & 71.95 & 66.70 & 69.07 & 71.95 & 65.03 & 68.30 & 83.66 &\bf 96.51 & 71.74 & 76.97 & 65.83 & 59.11 & 74.89 & 63.24 & 71.90 \\
  Bruce ICCV'17~\cite{Nie_2017_ICCV} & 90.1 & 88.2 & 85.7 & 95.6 & 103.9 & 90.4 & 117.9 & 136.4 & 98.5 & 103.0 & 92.4 & 94.4 & 90.6 & 86.0 & 89.5 & 97.5 \\
  Tekin ICCV'17~\cite{Tekin_2017_ICCV} & 53.91 & 62.19 & 61.51 & 66.18 & 80.12 & 64.61 & 83.17 &\bf 70.93 & 107.92 & 70.44 & 79.45 & 68.01 & 52.81 & 77.81 &63.11 & 70.81\\
  Zhou ICCV'17\cite{weak17ICCV} & 54.82 & 60.70 & 58.22 & 71.41 &\bf 62.03 & 53.83  & 55.58 & 75.20 & 111.59 & 64.15 & \bf 65.53& 66.05 & 63.22 &\bf 51.43  & 55.33 & 64.90 \\
  \textbf{Ours} & 50.03 & 59.96 & \bf54.66&\bf56.55&65.65&\bf52.74&\bf 54.81 & 85.85 & 117.98 &\bf62.48 &79.63&\bf  59.55&\bf 41.48&  65.21 &\bf48.52&\bf 63.67 \\
  \hline
  \hline
  \end{tabular}
  \end{adjustbox}

  \begin{center}
  \begin{tabular}{c}
  Protocol \#2 \\
  \end{tabular}
  \end{center}

  \begin{adjustbox}{max width=1.0\textwidth}
  \begin{tabular}{@{}l|ccccccccccccccc|c@{}}
  \hline
  \hline
   Method & Direction & Discuss & Eat & Greet & Phone & Pose & Purchase & Sit & SitDown & Smoke & Photo & Wait & Walk & WalkDog & WalkPair & Avg. \\
   \hline
  Akhter CVPR'15~\cite{pjal15CVPR} & 1199.20 & 177.60 & 161.80 & 197.80 & 176.20 & 195.40 & 167.30 & 160.70 & 173.70 & 177.80 & 186.50 & 181.90 & 198.60 & 176.20 & 192.70 & 181.56 \\
  Zhou PAMI'16 \cite{zhou16pami} & 99.70 & 95.80 & 87.90 & 116.80 & 108.30 & 93.50 & 95.30 & 109.10 & 137.50 & 106.00 & 107.30 & 102.20 & 110.40 & 106.50 & 115.20 & 106.10 \\
  Bogo ECCV'16 \cite{bogo2016keep} & 62.00 & 60.20 & 67.80 & 76.50 & 92.10 & 73.00 & 75.30 & 100.30 & 137.30 & 83.40 & 77.00 & 77.30 & 86.80 & 79.70 & 81.70 & 82.03 \\
  Moreno-Noguer CVPR'17\cite{DMR17CVPR} & 66.07 & 77.94 & 72.58 & 84.66 & 99.71 & 74.78 & 65.29 & 93.40 &\bf 103.14 & 85.03 & 98.52 & 98.78 & 78.12 & 80.05 & 74.77 & 83.52 \\
  Tome CVPR'17 \cite{lfd17CVPR} & - & - & - & - & - & - & - & - & - & - & - & - & - & - & - & 79.60 \\
  Chen CVPR'17 \cite{pem17CVPR} & 71.63 & 66.60 & 74.74 & 79.09 & 70.05 & 67.56 & 89.30 & 90.74 & 195.62 & 83.46 & 93.26 & 71.15 & 55.74 & 85.86 & 62.51 & 82.72  \\
  \textbf{Ours}& \bf 48.54&\bf 59.71& \bf56.12& \bf 56.12 &\bf67.68&\bf57.31&\bf55.57&\bf78.26  &115.85&\bf69.99&\bf71.47&\bf61.29  &\bf44.63&\bf62.22&\bf51.42&\bf63.74 \\
  \hline
  \hline
  \end{tabular}
  \end{adjustbox}

  \begin{center}
  \begin{tabular}{c}
  Protocol \#3 \\
  \end{tabular}
  \end{center}

  \begin{adjustbox}{max width=1.0\textwidth}
  \begin{tabular}{@{}l|ccccccccccccccc|c@{}}
  \hline
  \hline
   Method & Direction & Discuss & Eat & Greet & Phone & Pose & Purchase & Sit & SitDown & Smoke & Photo & Wait & Walk & WalkDog & WalkPair & Avg. \\
  \hline
    Yasin CVPR'16 \cite{ds16CVPR} & 88.40 & 72.50 & 108.50 & 110.20 & 97.10 & 81.60 & 107.20 & 119.00 & 170.80 & 108.20 & 142.50 & 86.90 & 92.10 & 165.70 & 102.00 & 110.18 \\
    Moreno-Noguer CVPR'17\cite{DMR17CVPR} & 67.44 & 63.76 & 87.15 & 73.91 & 71.48 & 69.88 & 65.08 & 71.69 & 98.63 & 81.33 & 93.25 & 74.62 & 76.51 & 77.72 & 74.63 & 76.47 \\
    Tome CVPR'17 \cite{lfd17CVPR} & - & - & - & - & - & - & - & - & - & - & - & - & - & - & - & 70.7 \\
    Bruce ICCV'17~\cite{Nie_2017_ICCV} & 62.8 & 69.2 & 79.6 & 78.8 & 80.8 & 72.5 & 73.9 & 96.1 & 106.9 & 88.0 & 86.9 & 70.7 & 71.9 & 76.5 & 73.2 &79.5 \\
  Sun ICCV'17\cite{Sun_2017_ICCV} & - & - &  - & - & - & - &  - & - & - & - & - &- &- &- &- & \bf 48.3 \\
\textbf{Ours} &\bf 38.69 &\bf 45.62 &\bf 54.77 &\bf 48.92 &\bf 54.65 &\bf 47.49 &\bf 47.17 &\bf 64.73 &\bf  94.30 &\bf  56.84 &\bf 78.85 &\bf 49.29 &\bf 33.07 &\bf 58.71 &\bf  38.96 &\bf 54.14 \\

\bottomrule
  \end{tabular}
  \end{adjustbox}

\end{table*}

\section{Experiments}
\label{sec:exper}
\subsection{Experimental Settings}
We perform extensive evaluations on two publicly available benchmarks: Human3.6M~\cite{huamn3.6m} and HumanEva-I~\cite{sigal2010humaneva}.

\textbf{Human3.6M dataset.} \kezen{The Human3.6M dataset} is a recently published dataset that provides 3.6 million 3D human pose images and corresponding annotations \kezen{from} a controlled laboratory environment. This dataset captures 11 professional actors performing in 15 scenarios under 4 different viewpoints. Moreover, there are three popular data partition protocols for this benchmark in the literature.
\begin{itemize}
\item {\em Protocol \#1:} The data from five subjects (S1, S5, S6, S7, and S8) are for training, and the data from two subjects (S9 and S11) are for testing. To increase the number of training samples, the sequences from different viewpoints of the same subject are treated as distinct sequences. \kezen{By} downsampling the frame rate from 50 FPS to 2 FPS, 62,437 human pose images (104 images per sequence) are obtained for training and 21,911 images are obtained for testing (91 images per sequence). This is the widely used evaluation protocol on Human3.6M, and it \kezen{was} followed by several works \cite{li2015maximum, Tekin_2016_CVPR, zhou2016deep, pem17CVPR}.
To be more general and make a fair comparison, our model is trained \kezen{both on} training samples from all 15 actions as previous works~\cite{li2015maximum, Tekin_2016_CVPR, zhou2016deep, pem17CVPR} and \kezen{by} exploiting individual \kezen{actions} as ~\cite{zhou2015sparseness,li2015maximum}.

\item {\em Protocol \#2:} \kezen{This protocol} only differs from Protocol \#1 in that only the frontal view is considered for testing, i.e., \kezen{testing is performed} on every 5-th frame of the sequences from the frontal camera (cam-3) from trial 1 of each activity with ground-truth cropping. The training data contain all actions and viewpoints.
\item {\em Protocol \#3:} Six subjects (S1, S5, S6, S7, S8 and S9) are used for training, and every 64-th frame of S11's video clips is used for testing. The training data contain all actions and viewpoints.
\end{itemize}

\textbf{HumanEva-I dataset.} \kezen{The HumanEva-I dataset} contains video sequences of four subjects performing six common actions (\eg, walking, jogging, boxing\kezen{, etc.}), and it also provides 3D pose annotation for each frame in the video sequences. We train our model on \kezen{the} training sequences of subjects 1, 2 and 3 and test on the `validation' sequence \kezen{under} the same protocol as \cite{yasin2016dual,Tekin_2016_CVPR,simo2012single,SimoSerraCVPR2013,BMVC2880,bo2010twin,radwan2013monocular,wang2014robust}.
Similar to Protocol \#1 of the Human3.6M dataset, the data from different camera viewpoints are also regarded as different training samples. Note that we \kezen{did not downsample} the video sequences to obtain more samples for training.

\begin{table*}[tp]
  \caption{Quantitative comparisons on the HumanEva-I dataset using 3D pose errors (in millimeters) for the ``walking'', ``jogging'' and ``boxing'' sequences. '-' indicates that the \kezen{author of the} corresponding method \kezen{did} not \kezen{report} the accuracy on that action. The entries with the smallest 3D pose errors for each category are bold-faced. Our model outperforms all the compared methods by a clear margin. }\label{table:hevai_res}
  \centering
  \scriptsize
  \resizebox{\textwidth}{!}{%
    \begin{tabular}{@{}l|cccc|cccc|cccc@{}}
      \toprule
      & \multicolumn{4}{c|}{Walking}                             & \multicolumn{4}{c|}{Jogging}                            & \multicolumn{4}{c}{Boxing}                                    \\
      Methods                  & S1            & S2            & S3            & Avg.       & S1            & S2            & S3            & Avg.       & S1            & S2            & S3            & Avg.       \\ \midrule
      Simo-Serra CVPR'12 \cite{simo2012single} & 99.6 & 108.3 & 127.4 & 111.8 & 109.2 & 93.1 & 115.8 & 108.9 & - & - & - & - \\
      Radwan ICCV'13 \cite{radwan2013monocular} & 75.1 & 99.8 & 93.8 & 89.6 & 79.2 & 89.8 & 99.4 & 89.5 & - & - & - & - \\
      Wang CVPR'14 \cite{wang2014robust} & 71.9 & 75.7 & 85.3 & 77.6 & 62.6 & 77.7 & 54.4 & 71.3 & - & - & - & - \\
      Du ECCV'16 \cite{DBLP:conf/eccv/DuWLHGWKG16} & 62.2 & 61.9 & 69.2 & 64.4 & 56.3 & 59.3 & 59.3 & 58.3 & - & - & - & - \\
      Simo-Serra CVPR'13 \cite{SimoSerraCVPR2013} & 65.1 & 48.6 & 73.5 & 62.4 & 74.2 & 46.6 & 32.2 & 56.7 & - & - & - & - \\
      Bo IJCV'10 \cite{bo2010twin}  & 45.4 & 28.3 & 62.3 & 45.3 & 55.1 & 43.2 & 37.4 & 45.2 & 42.5 & 64.0 & 69.3 & 58.6 \\
      Kostrikov BMVC'14 \cite{BMVC2880} & 44.0 & 30.9 & 41.7 & 38.9 & 57.2 & 35.0 & 33.3 & 40.3 & - & - & - & - \\
      Tekin CVPR'16 \cite{Tekin_2016_CVPR} & 37.5 & 25.1 & 49.2 & 37.3 & - & - & - & - & 50.5 & 61.7 & {57.5} & 56.6 \\
      Yasin CVPR'16 \cite{yasin2016dual} & 35.8 & 32.4 & 41.6 & 36.6 & 46.6 & 41.4 & 35.4 & 38.9 & - & - & - & - \\
      Lin CVPR'17\cite{rpsm17CVPR} & {26.5} & {20.7} & {38.0} & {28.4} & {41.0} & {29.7} & {29.1} & {33.2} & {39.4} & {57.8} & 61.2 & {52.8} \\
      Pavlakos CVPR'17~\cite{pavlakos2017volumetric} & 22.3 & 19.5 & 29.7 & 23.8 & 28.9 & 21.9 & 23.8 &  24.9 &- &- &- &- \\
      Moreno-Noguer CVPR'17 \cite{DMR17CVPR} & 19.7 & \bf 13.0 & 24.9 & 19.2 & 39.7 & 20.0 & 21.0 & 26.9 & - & - & - & - \\
      Tekin ICCV'17 \cite{Tekin_2017_ICCV} & 27.2 & 14.3 & 31.7 & 24.4 & - & - & - & - & - & - & - & - \\
      \bf Ours & \bf 17.2 & 13.4 & \bf20.5 & \bf17.0 &\bf 27.9 &  \bf19.5 &\bf 20.9 &\bf  22.8& \bf 29.7&\bf  44.0&\bf  47.2&\bf  40.3\\
      \bottomrule
    \end{tabular}%
  }  
\end{table*}

\begin{table}[t]
\vspace{-5pt}
\caption{Comparison of the average running time (milliseconds per image) on the Human3.6M benchmark. As shown in this table, our model performs nearly three times faster than the fastest of the compared methods. Specifically, the 2D pose sub-network costs 19ms, the 2D-to-3D pose transformer module costs 19 ms, and the 3D-to-2D pose projector module \kezen{costs} 13 ms.}\label{table:efficiency}
\vspace{-10pt}
\center
\small
\setlength{\tabcolsep}{2.5pt}
\begin{tabular}{|c|c|c|c|c|c|c|}
\hline
\hline
Method & Zhou  & Pavlakos & Zhou  & Tome & Ours  \\
& et al.~\cite{zhou2015sparseness} & et al.~\cite{pavlakos2017volumetric} & et al.~\cite{weak17ICCV} & et al.~\cite{lfd17CVPR}  & \\
\hline
Time & 880 & 174 & 311 & 444 & 51 \\
\hline
\hline
\end{tabular}
\vspace{-10pt}
\end{table}

\begin{figure*}[t]
\center
\includegraphics[width= 1 \textwidth]{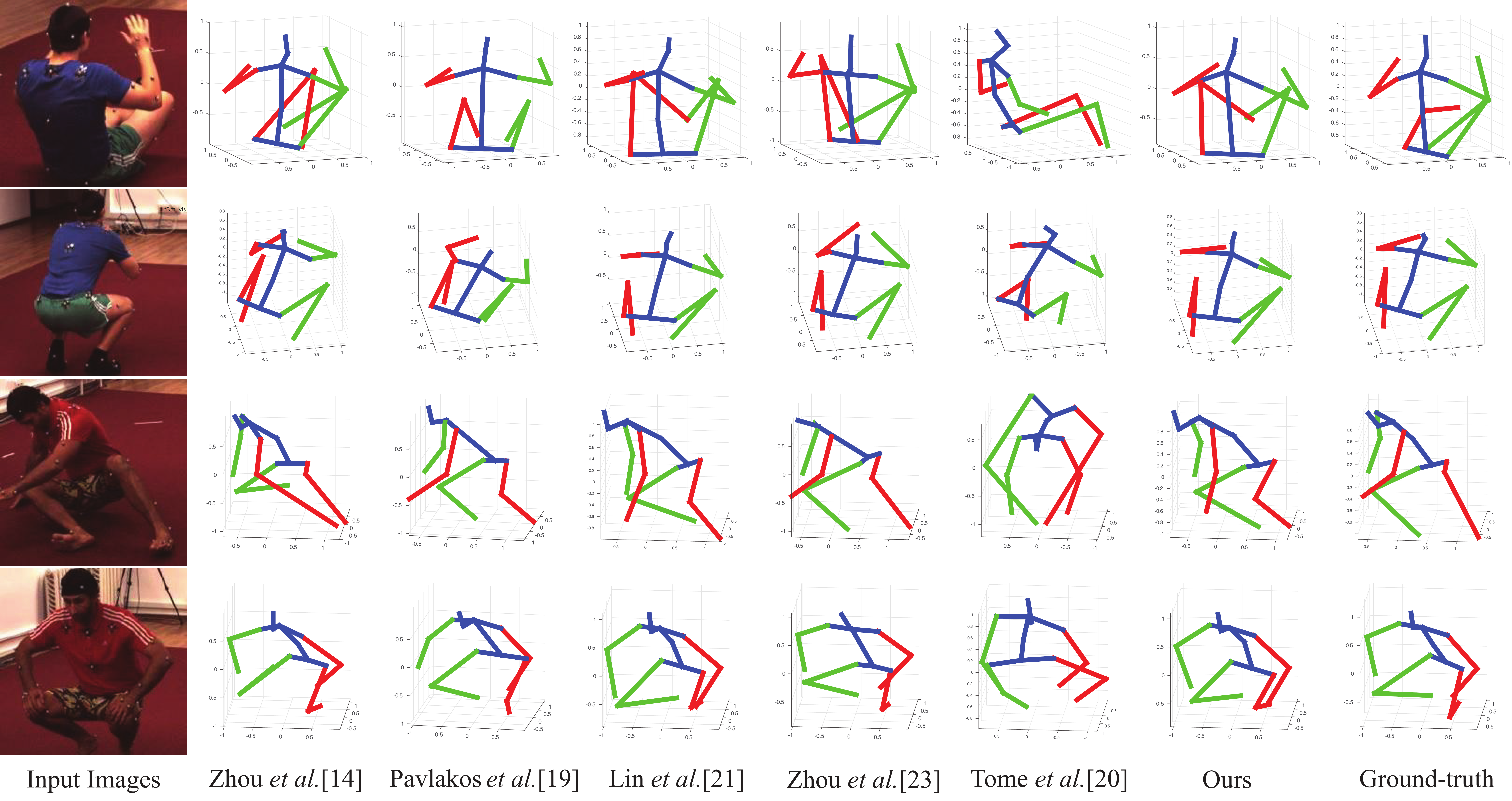}
\vspace{-15pt}
  \caption{\kezen{Qualitative} comparisons on the Human3.6M dataset. The 3D poses are visualized from the side view, and the cameras are depicted. The results from Zhou \etal~\cite{zhou2015sparseness}, Pavlakos \etal~\cite{pavlakos2017volumetric}, Lin \etal~\cite{rpsm17CVPR}, Zhou \etal~\cite{weak17ICCV}, Tome \etal~\cite{lfd17CVPR}, our model and the ground truth are illustrated from left to right. Our model achieves considerably more accurate estimations than all the compared methods. Best viewed in color. Red and green indicate left and right, respectively.}
  \vspace{-5pt}
  \label{fig:h3m_vis}
\end{figure*}

\begin{figure*}[t]
\center
\includegraphics[width= 0.7 \textwidth]{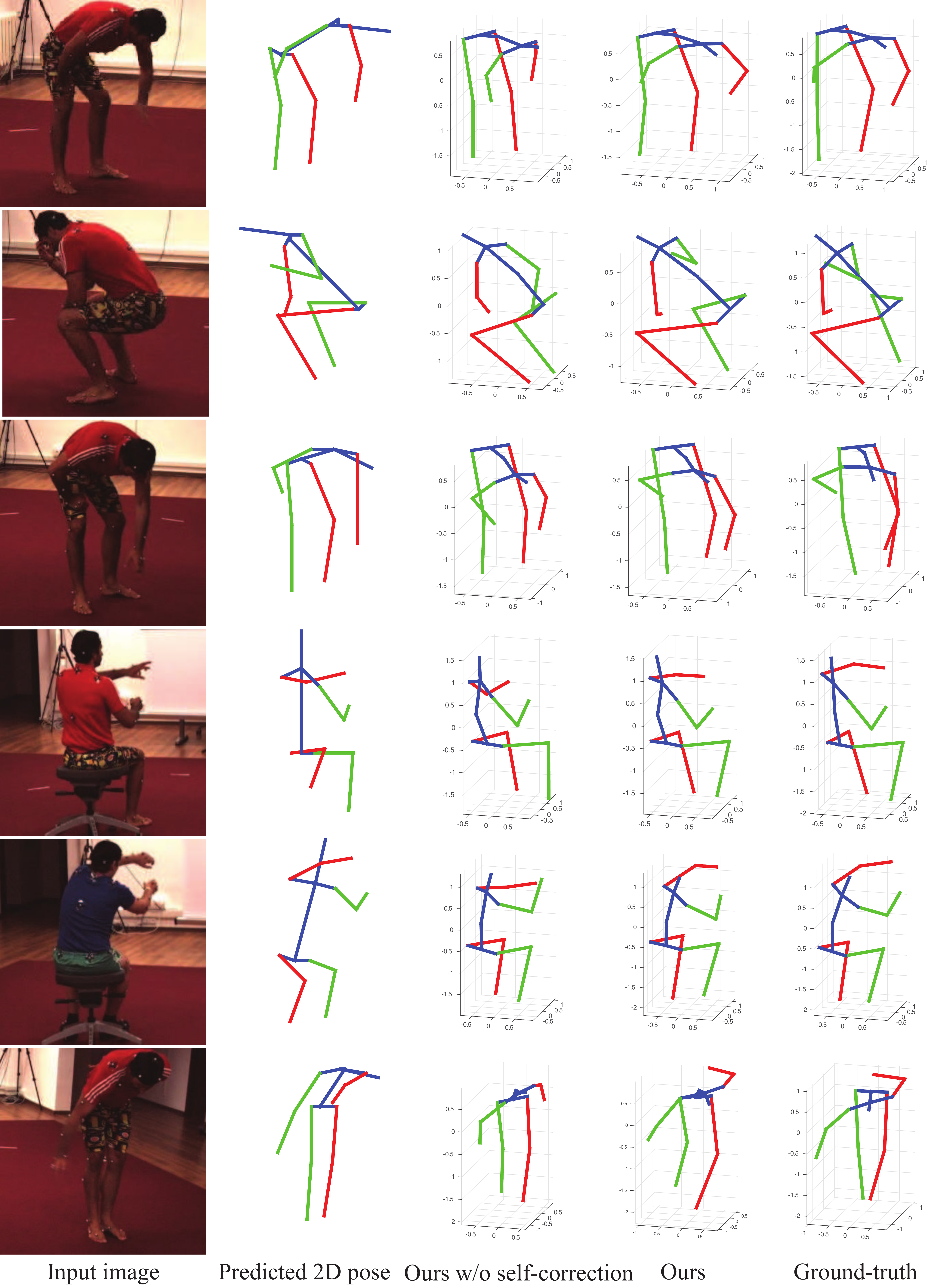}
  \vspace{-5pt}
  \caption{Qualitative comparisons of ours and ours w/o self-correction on the Human3.6M dataset. The input image, estimated 2D pose, ours w/o self-correction, ours and ground truth are listed from left to right, respectively. With the ground truth as reference, one can easily observe that the inaccurately predicted human 3D joints in ours w/o self-correction are effectively corrected in ours. Best viewed in color. Red and green indicate left and right, respectively.}
  \vspace{-8pt}
  \label{fig:ssr}
\end{figure*}

\begin{table*}[tpb]
  \caption{Empirical comparisons \kezen{under} different settings for ablation study using Protocol \#1. The entries with the smallest 3D pose errors on the Human3.6M dataset for each category are bold-faced.}
  \vspace{-5pt}
  \centering
  \setlength{\tabcolsep}{1.5pt}
  \begin{adjustbox}{max width=1.0\textwidth}
  \begin{tabular}{@{}c|ccccccccccccccc|c@{}}
  \toprule
  Method & Direction & Discuss & Eating & Greet & Phone & Pose & Purchase & Sitting & SitDown & Smoke & Photo & Wait & Walk & WalkDog & WalkPair & Avg. \\ \midrule
  {Ours w/o SSC train+test}  & 62.89 &  74.74&  67.86&  73.33 &79.76 &  67.48 &76.19& 100.21& 148.03 &  75.95&  100.26& 75.82&  58.03 & 78.74&  62.93&  80.15 \\
  \keze{Ours w/o SSC test}  &  \keze{52.95} & \keze{63.82} & \keze{57.15} & \keze{59.42} & \keze{68.83} & \keze{55.81} & \keze{57.75} & \keze{95.44} & \keze{125.70} & \keze{66.23} & \keze{82.91} & \keze{64.22} & \keze{44.24} & \keze{69.49} & \keze{50.54} & \keze{67.63}  \\
  \keze{Ours w/o projection} & \keze{69.46} & \keze{81.86} & \keze{74.46} & \keze{78.46} & \keze{85.14} & \keze{72.50} & \keze{85.39} & \keze{112.12} & \keze{158.14} & \keze{80.37} & \keze{115.46} & \keze{77.10} & \keze{58.38} & \keze{87.60 } & \keze{65.54} & \keze{86.80}  \\
  {Ours w/o temporal} & 70.46 & 83.36 & 76.46 & 80.96 & 88.14 & 76.00 & 92.39 & 116.62 & 163.14 & 85.87 & 111.46 & 83.60 & 65.38 & 95.10 & 73.54 & 90.83 \\
  {Ours w/ single frame} & 51.76 & 63.46 & 56.93 & 59.93 & 69.58 & 54.02 & 59.55 & 90.83 & 129.6 & 65.49 & 84.08 & 61.63 &45.07 & 70.33 & 53.21 & 67.70 \\
  {Ours w/o external} & 91.58 &109.35&93.28&98.52&102.16&93.87&118.15&134.94  &190.6&109.39&121.49&101.82&88.69&110.14&105.56&111.3 \\
  {Ours} &50.03& 59.96&54.66&56.55&65.65&52.74& 54.81&85.85&117.98&62.48	&79.63&	59.55&	41.48&	65.21	&48.52&	63.67 \\
  \keze{Ours w/ HG features} & \keze{49.34} & \keze{59.09} & \keze{54.08} & \keze{56.26} & \keze{64.48} & \keze{51.89} & \keze{54.09} & \keze{83.85} & \keze{116.55} & \keze{61.47} & \keze{78.52} & \keze{58.68} & \keze{41.48} & \keze{64.49} & \keze{48.48} & \keze{62.85} \\
  {Ours w/ 2D GT} & \bf  48.37& \bf 57.10& \bf 49.81& \bf 54.84& \bf 57.23& \bf 50.88& \bf 51.62& \bf 76.00& \bf 109.8	& \bf 55.28& \bf 74.52& \bf 56.98& \bf 40.16& \bf 61.29& \bf 47.15& \bf 59.41 \\
  \bottomrule
  \end{tabular}
  \end{adjustbox}
  \label{table:h3m_across_stages}
  \label{tab:rpsm-rho3-ext}
  \label{table:h3m_across_time}
\end{table*}

\begin{figure*}[t]
\center
\vspace{-10pt}
\includegraphics[width= 1 \textwidth]{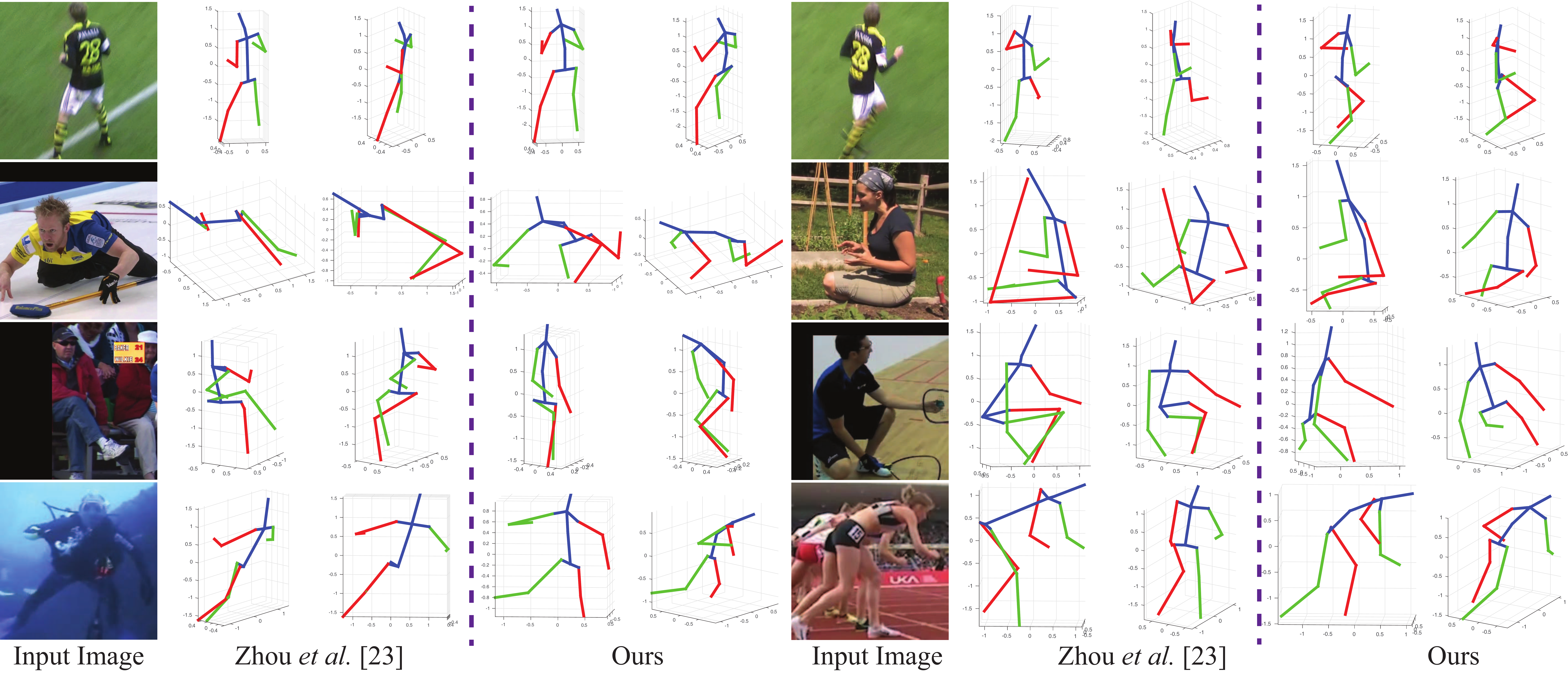}
  \vspace{-15pt}
  \caption{Some qualitative comparisons of our model and Zhou \etal~\cite{weak17ICCV} on two representative datasets in the wild, i.e., KTH Football II~\cite{kth_2013_BMVC} (first row) and MPII datasets~\cite{andriluka14cvpr} (the \kezen{remaining} rows). For each image, the original viewpoint and a better viewpoint are illustrated. Best viewed in color. Red and green indicate left and right, respectively.}
  \vspace{-8pt}
  \label{fig:2d_data}
\end{figure*}

\textbf{Implementation Details:}
\kezen{For} We follow \cite{graves2014towards} to build the LSTM memory cells, except that the peephole connections between cells and gates are omitted. Following~\cite{zhou2015sparseness,li2015maximum}, the input image is cropped around the human. To \kezen{maintain} the human width / height ratio, we crop a square image of the subject from the image according to the bounding box provided by the dataset. Then, we resize the image region inside the bounding box \kezen{to} 368$\times$368 before feeding it into our model. Moreover, we augment the training data \kezen{by simply performing} random scaling with factors in [0.9,1.1]. To transform the absolute locations of joint points into the [0,1] range, a $\text{max-min}$ normalization strategy is applied. In the testing phase, the predicted 3D pose is transformed to the origin scale according to the maximum and minimum values of the pose from the training frames. During the training, the Xavier initialization method \cite{glorot2010understanding} \kezen{is} used to initialize the weights of our model. \kezen{A} learning rate of $1e^{-5}$ is employed for training. The training phase \kezen{requires} approximately 2 days on a single NVIDIA GeForce GTX 1080.

\textbf{Evaluation metric.} Following \cite{zhou2015sparseness,DBLP:conf/eccv/DuWLHGWKG16,Tekin_2016_CVPR,rpsm17CVPR,weak17ICCV,pavlakos2017volumetric}, we employ the popular \emph{3D pose error} metric~\cite{simo2012single}, which calculates the Euclidean errors on all joints and all frames up to translation. In the following section, we report the 3D pose error metric for all the experimental comparisons and analyses.

\subsection{Comparisons with \kezen{Existing Methods}}
\textbf{Comparison on Human3.6M:}
\label{section:human3.6m_result}
We compare our model with the various \kezen{competing} methods on the Human3.6M \cite{huamn3.6m} and HumanEva-I \cite{sigal2010humaneva} datasets. For the fair comparison, we only consider the \kezen{competing} methods that do not need the intrinsic parameters of cameras for inference. This is reasonable for  practical use under various scenarios. These methods are LinKDE~\cite{huamn3.6m}, Tekin \etal \cite{Tekin_2016_CVPR}, Li \etal \cite{li2015maximum}, Zhou \etal \cite{zhou2015sparseness}, Zhou \etal \cite{zhou2016deep}, Du \etal \cite{DBLP:conf/eccv/DuWLHGWKG16}, Sanzari \etal \cite{DBLP:conf/eccv/SanzariNP16}, Yasin \etal~\cite{ds16CVPR}, and Bogo \etal~\cite{bogo2016keep}. Moreover, we compare \kezen{other competing} methods, i.e., Moreno-Noguer \etal~\cite{DMR17CVPR}, Tome \etal~\cite{lfd17CVPR}, Chen \etal~\cite{pem17CVPR}, Pavlakos \etal~\cite{pavlakos2017volumetric}, Zhou \etal~\cite{weak17ICCV}, Bruce \etal~\cite{Nie_2017_ICCV}, Tekin \etal~\cite{Tekin_2017_ICCV} and our conference version, i.e., Lin \etal~\cite{rpsm17CVPR}. For those compared methods (i.e., \cite{huamn3.6m, Tekin_2016_CVPR, li2015maximum, zhou2016deep, DBLP:conf/eccv/DuWLHGWKG16, DBLP:conf/eccv/SanzariNP16, ds16CVPR, bogo2016keep, DMR17CVPR, pem17CVPR, Nie_2017_ICCV}) whose source codes are not publicly available, we directly obtain their results from their published papers. For the other methods (i.e., \cite{zhou2015sparseness, lfd17CVPR, weak17ICCV, pavlakos2017volumetric, rpsm17CVPR, Tekin_2017_ICCV}), we directly use their official implementations for comparisons.



The results under three different protocols are summarized in Table \ref{table:h3m_res}. \kezen{Clearly,} our model outperforms all the competing methods (including \kezen{those} trained from the individual action as \kezen{in} \cite{pem17CVPR,li2015maximum,Tekin_2016_CVPR} and \kezen{on} all 15 actions) under {\em Protocol \#1}. Specifically, under the training from individual action setting, our model achieves superior performance on all the action types, and it \kezen{outperforms the best} competing methods with the joint mean error reduced by approximately 8\% compared with Tekin \etal~\cite{Tekin_2017_ICCV} (63.79mm {\em vs} 69.73mm). Under the training from all 15 actions, our model still consistently performs better than the compared approaches and obtains \kezen{better} accuracy. Notably, our model achieves a performance gain of nearly 12\% compared with our conference version (63.67mm {\em vs} 73.10 mm).

\kezen{The similar} superior performance of our model over all the compared methods can also be observed under {\em Protocol \#2} and {\em Protocol \#3}. Specifically, our model outperforms the best of the competing methods with the joint mean error reduced by approximately 19\% (63.74mm {\em vs} 79.6mm) under {\em Protocol \#2} and 16\% (54.14mm {\em vs} 70.70mm) under {\em Protocol \#3}.

In summary, our proposed model significantly outperforms all compared methods under all protocols with the mean error reduced by a clear margin. Note that some compared methods, \eg, \cite{li2015maximum,Tekin_2016_CVPR,DBLP:conf/eccv/DuWLHGWKG16,zhou2015sparseness,zhou2016deep,rpsm17CVPR,weak17ICCV,pavlakos2017volumetric,lfd17CVPR}, also employ deep learning techniques. In particular, Zhou \etal~\cite{zhou2016deep}'s method used the residual network~\cite{he2015deep}. Note that, the recently proposed methods all employ very deep network \kezen{architectures} (i.e.,  \cite{weak17ICCV} and \cite{pavlakos2017volumetric} proposed using Stacked Hourglass~\cite{newell2016stacked}, while \cite{rpsm17CVPR} and \cite{lfd17CVPR} employed CPM~\cite{wei2016convolutional}) to obtain satisfactory \kezen{accuracies}. This makes these methods time-consuming. In contrast to these methods, our model \kezen{achieves} a more lightweight architecture by replacing the multi-stage refinement in \cite{rpsm17CVPR} with the 3D-to-2D pose projector module. The superior performance achieved by our model demonstrates that our model is simple yet powerful in capturing complex contextual features within images, learning temporal dependencies within image sequences and preserving the geometric consistency within 3D pose predictions, which are critical for estimating 3D pose sequences. Some visual comparison results are presented in Fig.~\ref{fig:h3m_vis}.

\textbf{Comparison on HumanEva-I:}
We compare our model against \kezen{competing} methods, \kezen{including} discriminative regressions~\cite{bo2010twin,BMVC2880}, 2D pose detector-based methods \cite{simo2012single,SimoSerraCVPR2013,wang2014robust,yasin2016dual}, CNN-based approaches \cite{Tekin_2016_CVPR,yasin2016dual,pavlakos2017volumetric,DMR17CVPR,Tekin_2017_ICCV} and our preliminary version Lin \cite{rpsm17CVPR}, \kezen{on the HumanEva-I dataset}. For a fair comparison, our model also predicts the 3D pose consisting of 14 joints, i.e., left/right shoulder, elbow, wrist, left/right hip knee, ankle, head top and neck, as \cite{yasin2016dual}.

Table \ref{table:hevai_res} presents the performance comparisons of our model with all compared methods. \kezen{Clearly,} our model obtains substantially lower 3D pose errors than the compared methods on \keze{all} the \emph{walking, jogging and boxing} sequences. This result demonstrates the high generalization capability of our proposed model.

\subsection{Running \kezen{Time}}
\kezen{To compare} the efficiencies of our model and of the compared methods, we have conducted all the experiments on a desktop with an Intel 3.4GHz CPU and a single NVIDIA GeForce GTX 1080 \kezen{GPU}. In terms of time efficiency, compared with \cite{zhou2015sparseness} (880 ms per image), \cite{pavlakos2017volumetric} (170 ms per image), \cite{weak17ICCV}  (311 ms per image), and \cite{lfd17CVPR} (444 ms per image), our model model only \kezen{requires} 51 ms per image. The detailed time analysis is presented in Table~\ref{table:efficiency}. Our model performs approximately 3 times faster than \cite{pavlakos2017volumetric}, which is the fastest of the compared methods. Moreover, although only performing slightly better than the best of the compared methods \cite{weak17ICCV} under {\em Protocol \#1}, our model runs nearly 6 times faster, thanks to the 3D-to-2D pose projector module \kezen{enabling a} lightweight architecture. This result validates the efficiency of our proposed model.

\subsection{\keze{Ablation Study}}
To perform a detailed component analysis, we conducted the experiments on the Human3.6M \keze{benchmark} under the {\em Protocol \#1} and our proposed model was trained on all \keze{actions} for a fair comparison.

\subsubsection{\keze{Self-supervised Correction} Mechanism}
\keze{To demonstrate the superiority of the proposed \keze{self-supervised} correction (SSC) mechanism, we conduct the following experiment: disabling this module in both training and inference phase by directly regarding the intermediate 3D pose predictions as the final output \keze{(denoted as ``ours w/o SSC train+test'')}. Moreover, we have also disabled the self-supervised correction mechanism during the inference phase and denote this version as “ours w/o SSC test".

The results in Table~\ref{tab:rpsm-rho3-ext} demonstrate that our w/o SSC test significantly outperforms ours w/o SSC train+test (67.63mm vs 80.15mm), and our model surpasses ours w/o SSC test by a clear margin (63.67mm vs. 67.63mm). These observations justify the contribution of the proposed SSC mechanism. This result demonstrates that the \keze{self-supervised correction} mechanism is highly beneficial for improving the performance both in the training and testing phase.} Qualitative comparison results on the Human3.6M dataset are shown in Fig.~\ref{fig:ssr}. As depicted in Fig.~\ref{fig:ssr}, the intermediate 3D pose predictions (i.e., ours w/o self-correction) contain several inaccurate joint locations compared with the \kezen{ground truth} because of the self-occlusion of body parts. However, the predicted 2D poses are of high accuracy \kezen{because of} the powerful CNN, which is trained from a large variety of 2D pose data. Because the estimated 2D poses are more reliable, our proposed 3D-to-2D pose projector module can utilize them as optimization objectives to \keze{bidirectionally} refine the 3D pose predictions. As shown in Fig.~\ref{fig:ssr}, the joint predictions are effectively corrected by our model. This experiment clearly demonstrates that our 3D-to-2D pose projector, utilizing these predicted 2D poses as guidance, can contribute to enhancing 3D human pose estimation by correcting the 3D pose predictions without additional 3D \keze{pose} annotations.

 To further \kezen{upper bound the} performance of the proposed \keze{self-supervised correction} mechanism \keze{in the testing phase}, we directly \keze{employ the 2D pose ground-truth }to correct the intermediate predicted 3D poses. We denote this version of our model as ``ours w/ 2D GT''. Table~\ref{tab:rpsm-rho3-ext} demonstrates that ours w/ 2D GT achieves significantly better results than our model by reducing the mean joint error by approximately 6\% (59.41mm {\em vs} 63.67mm). \keze{Moreover, we have also implemented another 2D pose prediction model from the hourglass network with two stages for self-supervised correction (denoted as ``ours w/ HG features''). Our method w/ HG features performs slightly better than ours (62.85mm vs 63.67mm).} This result \kezen{evidences} the effectiveness of our designed 3D-to-2D pose projector module in \keze{bidirectionally} refining the intermediate 3D pose estimation.

\keze{
We have further compared our method with a multi-task approach that simultaneously estimates 3D and 2D pose without re-projection (denoted as ``ours w/o projection''). Specifically, ours w/o projection shares the same network architecture as our full model. The only difference is that ours w/o projection directly estimates both 2D and 3D poses during the training phase. Since it is quite difficult to directly train the network into convergence with huge 2D pose data and relatively small 3D pose data, we first optimize the network parameters by using the 2D pose data only from the MPII dataset, and further perform training with the 2D pose data and 3D pose data from the Human3.6M dataset.

As illustrated in Table~\ref{tab:rpsm-rho3-ext}, our full model performs substantially better than ours w/o projection (63.67mm vs 86.80mm). The reason may be that directly regressing the 2D pose may mislead the main learning task of the model, i.e., the network concentrates on improving the overall performance of both 2D and 3D pose prediction. This demonstrates the effectiveness of the proposed 3D-to-2D pose projector module.
}

\subsubsection{Temporal Dependency}
The model performance without temporal dependency in the training phase is also compared in Table~\ref{tab:rpsm-rho3-ext} (denoted as ``ours w/o temporal''). Note that the input of ours w/o temporal is only a single image rather than a sequence \kezen{in} the training and testing \kezen{phases}. Hence, the temporal information is ignored. Thus, the LSTM layer \kezen{for the} 3D pose errors is replaced \kezen{by} a fully connected layer with the same units as the LSTM layer. As illustrated in Table~\ref{tab:rpsm-rho3-ext}, ours w/o temporal has \kezen{suffered} considerably higher 3D pose errors than ours (90.83mm vs 63.67mm). Moreover, we have also analyzed the contribution of the temporal dependency in the testing phase. To discard the temporal dependency during the inference \keze{phase}, we \kezen{regarded} a single frame as input to evaluate the performance of our model, and we denote it as ``ours w/ single frame''. Table~\ref{tab:rpsm-rho3-ext} demonstrates that \keze{this variant} performs worse \kezen{compared to} ours by increasing the mean joint error by approximately 6\% (67.70mm {\em vs} 63.67mm). This result validates the contribution of temporal information for 3D human pose estimation during the training and testing \kezen{phases}.

\subsubsection{External 2D Human Pose Data}
To evaluate the performance without external 2D human pose data, we have only employed 3D pose data with 3D and 2D annotations from Human3.6M for training our model. We denote this version of our model as ``ours w/o external''. As shown in Table~\ref{tab:rpsm-rho3-ext}, \kezen{the} ours w/o external version performs quite worse than our model (63.67mm {\em vs} 111.30mm). This reason may be \kezen{that} the training samples \keze{from Human3.6M}, compared with the MPII dataset~\cite{andriluka14cvpr}, are less challenging\kezen{, therein having} fewer variations for our model to learn a rich and powerful 2D pose presentation. Thanks to the proposed self-supervised correction mechanism, our model can effectively leverage a large variety of 2D human pose data to improve the performance of the 3D human pose estimation.

Moreover, \keze{in terms of} estimating 3D human pose in the wild, our model advances the existing method~\cite{weak17ICCV} in leveraging more abundant 3D geometric knowledge for mining  in-the-wild 2D human pose data. Instead of oversimplifying the 3D geometric constraint as the relative bone length in \cite{weak17ICCV}, our model introduces a \keze{self-supervised correction mechanism} to retain the 3D geometric consistency between the 2D projections of 3D poses and the estimated 2D poses. Therefore, our model can \keze{bidirectionally} refine the 3D pose predictions in a self-supervised manner. Fig.~\ref{fig:2d_data} presents qualitative comparisons for images taken from the KTH Football II~\cite{kth_2013_BMVC} and MPII dataset~\cite{andriluka14cvpr}, \keze{respectively}. As \keze{shown in} Fig.~\ref{fig:2d_data}, our model achieves 3D pose predictions of \kezen{superior accuracy compared to the competing} method~\cite{weak17ICCV}.

\section{Conclusion}
\label{sec:conclude}
This paper presented a 3D human \kezen{pose machine that} can learn to integrate rich spatio-temporal long-range dependencies and 3D geometry knowledge in an implicit and comprehensive manner. We further enhanced our model by developing a novel self-supervised correction mechanism, which involves two dual learning tasks, i.e., 2D-to-3D pose transformation and 3D-to-2D pose projection, under a \keze{self-supervised correction} mechanism. This mechanism retains the geometric consistency between the 2D projections of 3D poses and the estimated 2D poses, and it enables our model to utilize the estimated 2D human pose to \keze{bidirectionally} refine the intermediate 3D pose estimation. Therefore, our proposed self-supervised correction mechanism can bridge the domain gap between 3D and 2D human poses to leverage the external 2D human pose data without requiring additional 3D annotations. \kezen{Extensive} evaluations on two public 3D human pose datasets validate the effectiveness and \kezen{superiority} of our proposed model. In \keze{future work}, focusing on sequence-based human centric analyses (e.g., human action and activity recognition), we will extend our proposed \keze{self-supervised correction} mechanism for temporal relationship modeling, and design new self-supervision objectives to incorporating abundant 3D geometric knowledge for training models in a cost-effective \kezen{manner}.


%

%
%

\ifCLASSOPTIONcompsoc
  \section*{Acknowledgments}
\else
  \section*{Acknowledgment}
\fi

This work was supported in part by the National Key Research and Development Program of China under Grant No. 2018YFC0830103, in part by National High Level Talents Special Support Plan (Ten Thousand Talents Program), in part by National Natural Science Foundation of China (NSFC) under Grant No. 61622214, and 61836012, in part by the Ministry of Public Security Science and Technology Police Foundation Project No. 2016GABJC48, and in part by Guandong ``Climbing Program'' special funds under Grant pdjh2018b0013.

\ifCLASSOPTIONcaptionsoff
  \newpage
\fi



%

%

\bibliographystyle{IEEEtran}

\bibliography{human_pose}

\begin{IEEEbiography}[{\includegraphics[width=1in,height=1.25in,clip,keepaspectratio]{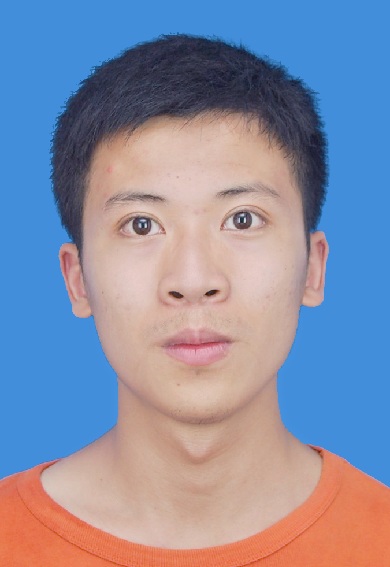}}]{Keze Wang} received his B.S. degree in software engineering from Sun Yat-Sen University, Guangzhou, China, in 2012. He is currently pursuing his dual Ph.D. degree at Sun Yat-Sen University and Hong Kong Polytechnic University, advised by Prof. Liang Lin and Lei Zhang. His current research interests include computer vision and machine learning. More information can be found on his personal website \url{http://kezewang.com}.
\end{IEEEbiography}

\begin{IEEEbiography}[{\includegraphics[width=1in,height=1.25in,clip,keepaspectratio]{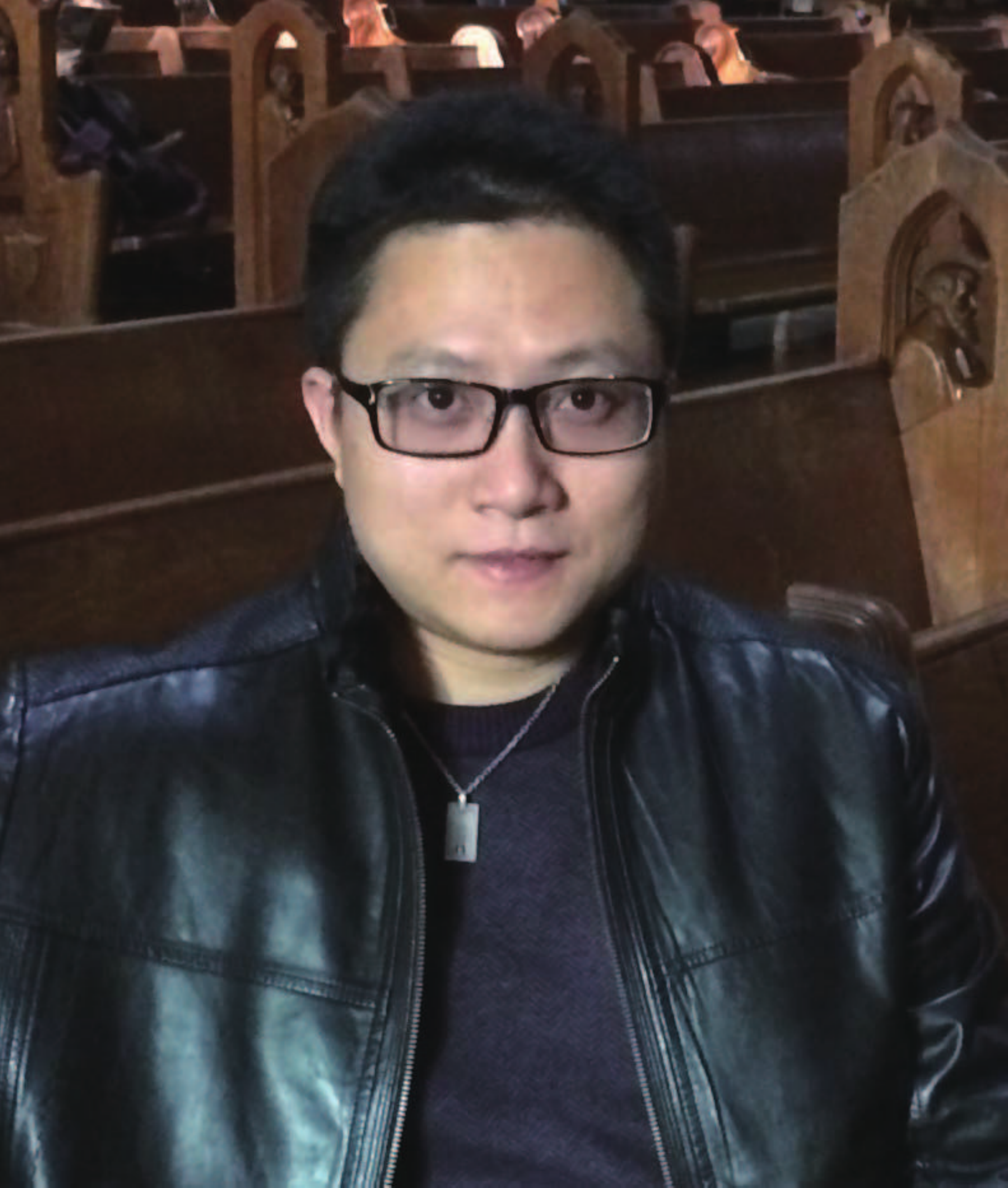}}]{Liang Lin}, IET Fellow, a full Professor of Sun Yat-sen University. From 2008 to 2010, he was a Post-Doctoral Fellow at University of California, Los Angeles. He currently leads the SenseTime R\&D teams in developing cutting-edge and deliverable solutions in computer vision, data analysis and mining, and intelligent robotic systems. He has authored and co-authored more than 100 papers in top-tier academic journals and conferences. He is currently serving as an associate editor of IEEE Trans. Human-Machine Systems. He served as Area/Session Chairs for numerous conferences such as CVPR, ICME, ACCV, and ICMR. He was the recipient of Best Paper Runners-Up Award at ACM NPAR 2010, the Google Faculty Award in 2012, the Best Paper Diamond Award at IEEE ICME 2017, and the Hong Kong Scholars Award in 2014.
\end{IEEEbiography}

\begin{IEEEbiography}[{\includegraphics[width=1in,height=1.25in,clip,keepaspectratio]{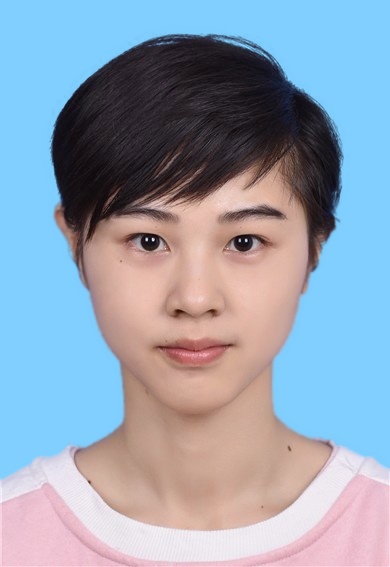}}]{Chenhan Jiang} received her BS degree in Software Engineering from XiDian University, Xi'an, China. She is currently pursuing her Master's Degree at the School of Data and Computer Science at Sun Yat-Sen University, advised by Professor Liang Lin. Her current research interests include computer vision (e.g., 3D human pose estimation) and machine learning.
\end{IEEEbiography}

\begin{IEEEbiography}[{\includegraphics[width=1in,height=1.25in,clip,keepaspectratio]{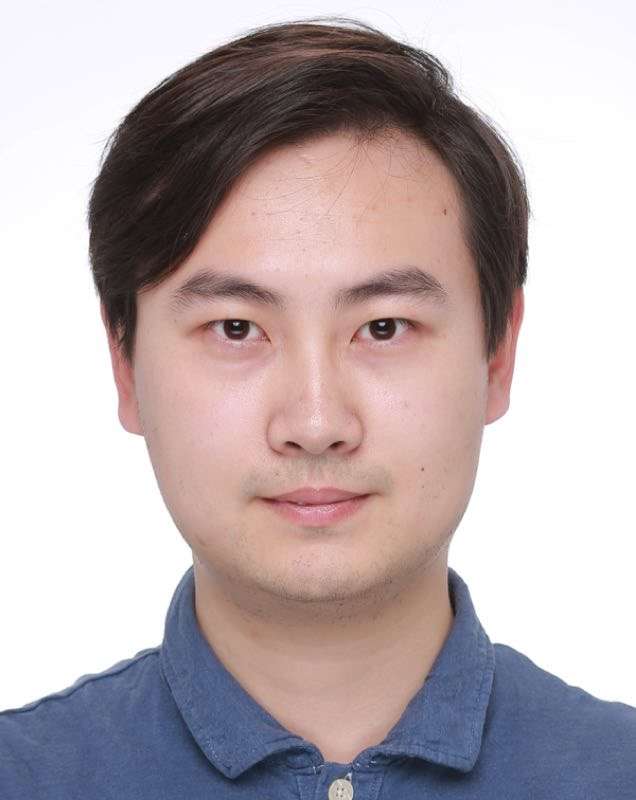}}]{Chen Qian} received his M.Phil. degree from the Department of Information Engineering, the Chinese University of Hong Kong, in 2014, and his B.Eng. degree from the Institute for Interdisciplinary Information Science, Tsinghua University, in 2012. He is currently working at SenseTime as research director. His research interests include human-related computer vision and machine learning problems.
\end{IEEEbiography}

\begin{IEEEbiography}[{\includegraphics[width=1in,height=1.25in,clip,keepaspectratio]{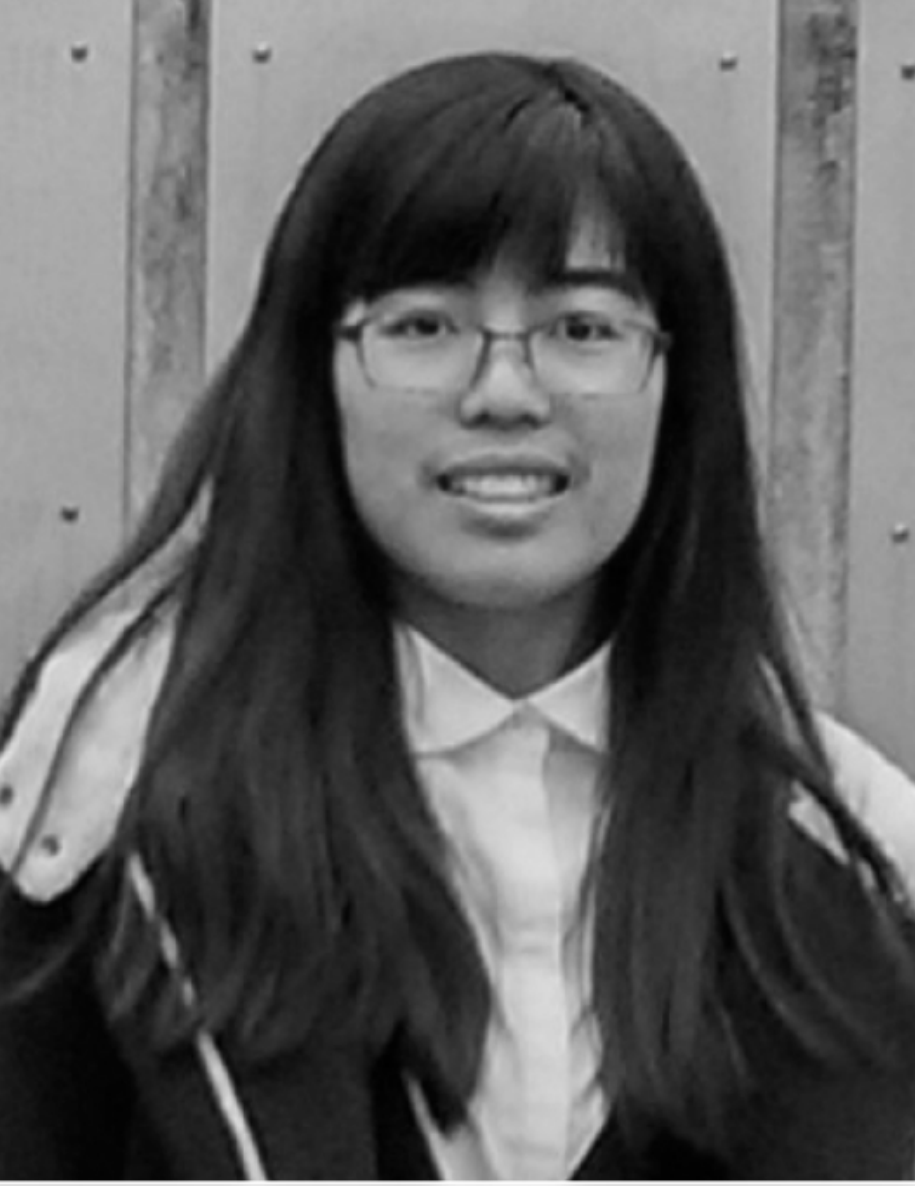}}]{Pengxu Wei}received the B.S. degree in computer science and technology from the China University of Mining and Technology, Beijing, China, in 2011 and the Ph.D. degree with the School of Electronic, Electrical, and Communication Engineering, University of Chinese Academy of Sciences in 2018. Her current research interests include computer vision and machine learning, specifically for data-driven vision and scene image recognition. She is currently a research scientist at Sun Yat-sen University.
\end{IEEEbiography}


\vfill


\end{document}